\documentclass[pdflatex,sn-mathphys]{sn-jnl}% Math and Physical Sciences Reference Style
\jyear{2021}%

\theoremstyle{thmstyleone}%
\theoremstyle{thmstyletwo}%

\theoremstyle{thmstylethree}%

\usepackage{subcaption}
\newcommand{\specialcell}[2][c]{%
    \begin{tabular}[#1]{@{}c@{}}#2\end{tabular}
}
  
\begin{document}

\title[Personalised Meta-path Generation for Heterogeneous GNNs]{Personalised Meta-path Generation for Heterogeneous Graph Neural Networks}


\author[1]{\fnm{Zhiqiang} \sur{Zhong}}\email{zhiqiang.zhong@@uni.lu}

\author*[2]{\fnm{Cheng-Te} \sur{Li}}\email{chengte@mail.ncku.edu.tw}

\author*[1,3]{\fnm{Jun} \sur{Pang}}\email{jun.pang@uni.lu}

\affil[1]{\orgdiv{Faculty of Science, Technology and Medicine}, \orgname{University of Luxembourg}, \orgaddress{\city{Esch-sur-Alzette}, \country{Luxembourg}}}

\affil[2]{\orgdiv{Institute of Data Science and the Department of Statistics}, \orgname{National Cheng Kung University}, \orgaddress{\city{Tainan}, \country{Taiwan}}}

\affil[3]{\orgdiv{Interdisciplinary Centre for Security, Reliability and Trust}, \orgname{University of Luxembourg}, \orgaddress{\city{Esch-sur-Alzette}, \country{Luxembourg}}}


% \abstract{The abstract serves both as a general introduction to the topic and as a brief, non-technical summary of the main results and their implications. Authors are advised to check the author instructions for the journal they are submitting to for word limits and if structural elements like subheadings, citations, or equations are permitted.}

%------------------------------------------------------------------------------
\abstract{
Recently, increasing attention has been paid to heterogeneous graph representation learning (HGRL), which aims to embed rich structural and semantic information in heterogeneous information networks (HINs) into low-dimensional node representations.
To date, most HGRL models rely on hand-crafted meta-paths.
However, the dependency on manually-defined meta-paths requires domain knowledge, which is difficult to obtain for complex HINs.
More importantly, the pre-defined or generated meta-paths of all existing HGRL methods attached to each node type or node pair cannot be personalised to each individual node.
To fully unleash the power of HGRL, we present a novel framework, Personalised Meta-path based Heterogeneous Graph Neural Networks (PM-HGNN), to jointly generate meta-paths that are personalised for each individual node in a HIN and learn node representations for the target downstream task like node classification.
Precisely, PM-HGNN treats the meta-path generation as a Markov Decision Process and utilises a policy network to adaptively generate a meta-path for each individual node and simultaneously learn effective node representations.
The policy network is trained with deep reinforcement learning by exploiting the performance improvement on a downstream task.
We further propose an extension, PM-HGNN\textit{++}, to better encode relational structure and accelerate the training during the meta-path generation.
Experimental results reveal that both PM-HGNN and PM-HGNN\textit{++} can significantly and consistently outperform $16$ competing baselines and state-of-the-art methods in various settings of node classification. 
Qualitative analysis also shows that PM-HGNN\textit{++} can identify meaningful meta-paths overlooked by human knowledge.

}
%------------------------------------------------------------------------------

\keywords{Meta-path generation, heterogeneous graph neural networks}



\maketitle

%------------------------------------------------------------------------------
\section{Introduction}
\label{sec:introduction}
\begin{figure}[ht!]
\centering
\includegraphics[width=.6\linewidth]{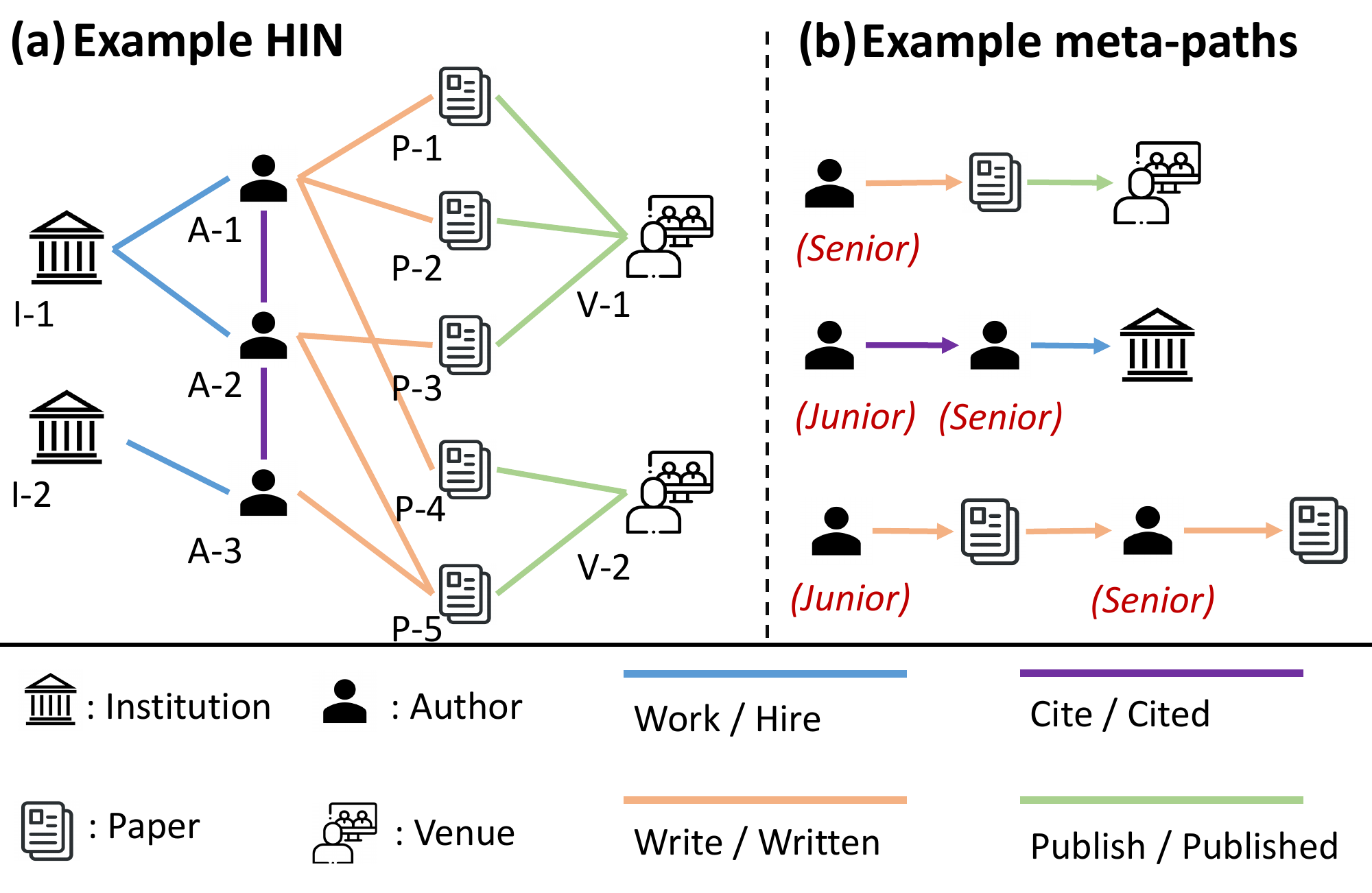}
\caption{
    An example HIN and few meth-paths:
    (a) an academia HIN;
    (b) meta-paths designed for {\it senior} vs. {\it junior} authors. 
}
\label{fig:example_HIN}
\vspace{-3mm}
\end{figure}

The complex interactions in real-world data, such as social networks, biological networks and knowledge graphs, can be modelled as Heterogeneous Information Networks (HINs)~\cite{SH12}, which are commonly associated with multiple types of nodes and relations.
Take the academia HIN depicted in Fig.~\ref{fig:example_HIN}-(a) as an example; it involves $4$ types of nodes, including {\it Papers} (\textit{P}), {\it Authors} (\textit{A}), {\it Institutions} (\textit{I}) and {\it publication venues} (\textit{V}), and $8$ types of relations.
Due to the capability of HINs to depict the complex inter-dependency between nodes, they have attracted increasing attention in the research community and have been applied in the fields of relational learning~\cite{V05}, recommender systems~\cite{ZMGZ21}, information retrieval~\cite{WDPH20}, etc.
However, the complex semantics and non-Euclidean nature of HINs render them challenging to modulate by conventional machine learning algorithms designed for tabular or sequential data.

Over the past decade, a significant line of research on HINs is \textit{Heterogeneous Graph Representation Learning} (HGRL).
The goal of HGRL is to learn latent node representations, which encode complex graph structure and multi-typed nodes, for downstream tasks, including link prediction~\cite{WDPH20}, node classification~\cite{WJSWYCY19} and node clustering~\cite{FZMK20}.
As discussed in a recent survey~\cite{DHWST20}, one of the paradigms on HGRL is to manually define and use meta-paths to model HIN's rich semantics and to leverage random walks to transform the graph structure into a set of sequences~\cite{DCS17,FLL17,SHZY19}, which can be further exploited by shallow embedding learning algorithms~\cite{MSCCD13,LM14}.
A meta-path scheme is defined as a sequence of relations over HIN's network schema.
For instance, an illustrative meta-path in the academia HIN in Fig.~\ref{fig:example_HIN}-(a) is 
\textit{A}$\xrightarrow{Cite}$\textit{A}$\xrightarrow{\it Write}$\textit{P}.
Some follow-up shallow HGRL models try to avoid the requirement of manually defined meta-paths by developing jump and stay random walk strategies~\cite{HYC18}, performing random walk with the guide of node contexts~\cite{JLJW20}, or switching to utilising network schema~\cite{ZWSLY20}.
Nevertheless, these ``shallow'' methods neither support end-to-end training to learn more effective representations for a specific task nor fully utilise node attributes due to the limitation of the embedding algorithms.

Recently, in view of the impressive success of Graph Neural Networks (GNNs)~\cite{KW17,VCCRLB18}, the second paradigm of HGRL attempts to devise Heterogeneous Graph Neural Networks (HGNNs) for HGRL~\cite{SKBBTW18,ZSHSC19,FZMK20}, which extend various graph convolutions on HIN.
Compared with ``shallow'' HGRL methods, HGNNs support an end-to-end training mechanism that can learn node representations with some labelled nodes' assistance and are also empowered by more complex encoders instead of using the shallow embedding learning methods. 
HGNNs can model both structure and node attributes in HINs with the guidance of meta-paths.
However, they still rely on the hand-crafted meta-paths to explicitly model the semantics of HINs, and obtaining meaningful and useful meta-paths for nodes in HINs to guide HGNNs is still highly non-trivial. 

More precisely, existing meta-path guided HGRL methods simply assume that nodes with the same type share the same meta-path.
Take the academia HIN as an example (Fig.~\ref{fig:example_HIN}-(a)) and assume we plan to learn node representations to determine the research area of ${\it Author}$s.
A meta-path $\Omega_{1}$: \textit{A}$\xrightarrow{\it Write}$\textit{P}$\xrightarrow{\it Published}$\textit{V} may be useful to learn a representation of a \textit{senior} researcher since his/her published papers and attended venues may provide sufficient information to decide his/her research area.
While learning the representation of a \textit{junior} PhD candidate with just a few published papers, we may need to extract information from his collaborators following the meta-path: $\Omega_{2}$: \textit{A}$\xrightarrow{\it Cite}$\textit{A}$\xrightarrow{\it Write}$\textit{P}, because $\Omega_{1}$ retains little information in the case of junior PhD candidates.
Hence, we argue that we should generate a \textit{personalised} meta-path for each ``individual node'' according to its attributes and neighbouring relational structures instead of giving each ``node type'' several pre-defined meta-paths in general. 

Motivated by the outstanding success of Reinforcement Learning (RL) in strategy selection problems~\cite{MKSRVBGRF15}, previous methods attempt to apply RL techniques to find paths between given node pairs which model the similarity between two nodes~\cite{MCMSZ15,YLHZPH18,WDPH20}. 
The found paths are then fed into the encoder to learn representations for pairwise tasks like link prediction.
Nevertheless, challenges still remain in designing personalised meta-paths of individual nodes for node-wise tasks. 

\smallskip\noindent
{\bf Key Challenges in Personalised Meta-paths Generation}. 
First, the definition of meta-paths requires rich domain knowledge that is extremely difficult to obtain in complex and semantic rich HINs~\cite{DHWST20}.
Specifically, given a HIN $G$ with a node type set $N$, a relation type set $R$ and a fixed meta-path length $T$.
The possible meta-paths are contained in a set of size $(|N| \times |R|)^{T}$.
Such a huge set can result in a combinatorial explosion when increasing the scales of $|N|$, $|R|$ and $T$.
Second, the representation capacities of manually-defined meta-paths are limited to a specific task on specific HIN since different $G$ with the same $N$ and $R$ may have different node attributes and relation types distributions.
It requires defining appropriate meta-paths for each task on each HIN, which is extremely difficult for practical applications.

In light of these challenges, we propose to investigate HGRL with the objectives of (i) learning to generate a personalised meta-path for each individual node in HINs automatically, (ii) learning node representations effectively and efficiently with personalised meta-paths, and (iii) retaining the end-to-end training strategy to achieve task-oriented optimisation.
To achieve these objectives, we present a novel \textbf{P}ersonalised \textbf{M}eta-path based \textbf{H}eterogeneous \textbf{G}raph \textbf{N}eural \textbf{N}etwork (PM-HGNN), to unleash the power of HGRL.

\smallskip\noindent
{\bf Key Ideas of PM-HGNN}.
Generally, we aim to replace the human efforts on meta-path generation with an RL agent to address the limitations of the dependency on hand-crafted meta-paths of HGNNs.
Compared with experts with domain knowledge, the RL agent can adaptively generate personalised meta-paths for each individual node in terms of a specific task/HIN through sequential exploration and exploitation. 
That said, the obtained meta-paths are no longer for specific types of nodes but are personalised for each individual node. 
Both graph structure and node attributes are considered in the meta-path generation process, and it is practicable for HINs with complex semantics.

As illustrated in Fig.~\ref{fig:metapath_gene_mdp}, the meta-path generation process can be naturally considered as a Markov Decision Process (MDP), in which the next relation to extending the meta-path depends on the current state in a meta-path.
Moreover, an HGNN model is proposed to learn node representations from the derived meta-paths that can be applied to a downstream task, such as node classification.
We propose to employ a policy network (agent) to solve this MDP and use an RL algorithm enhanced by a reward function to train the agent.
The reward function is defined as the performance improvement over historical performance, which encourages the RL agent to achieve better and more stable performance. 
In addition, we find that there exists a large computational redundancy during the information aggregation on meta-paths, thus we develop an efficient strategy to achieve redundancy-free aggregation.

We showcase an instance of our framework, i.e. PM-HGNN, by implementing it with a classic RL algorithm, i.e., Deep Q-learning~\cite{MKSRVBGRF15}. 
Besides, we further propose an extension of PM-HGNN\textit{++} to deal with the issues of high computational cost and ignoring relational information in PM-HGNN. 
Specifically, PM-HGNN generates a meta-path for each node according to node attributes, while PM-HGNN\textit{++} further enables the meta-path generation to explore the structural semantics of HIN.
PM-HGNN\textit{++} is able to not only significantly accelerate the HGRL process but also improve the effectiveness of learned node representations with promising performance on downstream tasks.

\smallskip\noindent
{\bf Main Contributions}.
We summarise our contributions below:
\begin{itemize}
    \item We present a framework, PM-HGNN\footnote{Code and data are available at: \url{https://github.com/zhiqiangzhongddu/PM-HGNN}}, to learn node representations in a HIN without hand-crafted meta-paths. An essential novelty of PM-HGNN is that the generated meta-paths are personalised to every individual node rather than general to each node type.
    \item We propose an attention-based redundancy-free mechanism to reduce redundant computation during heterogeneous information aggregation on the derived meta-path instances.
    \item We further develop an extension of PM-HGNN, PM-HGNN\textit{++}, which not only improves the meta-path generation by incorporating node attributes and relational structure but also accelerates the training process.
    \item Experiments conducted on node classification tasks with unsupervised and (semi-)supervised settings exhibit that our framework can significantly and consistently outperform $16$ competing methods (up to $5.6\%$ {\it Micro-F1} improvements).
    Advanced studies further reveal that PM-HGNN\textit{++} can identify meaningful meta-paths that human experts have ignored. 
\end{itemize}

%------------------------------------------------------------------------------

%------------------------------------------------------------------------------
\section{Related work} % (fold)
\label{sec:related_work}
\textbf{Relational Learning.}
In the past decades, research focused on using frameworks that could represent a variable number of entities and the relationships that hold amongst them. 
The interest in learning using this expressive representation formalism soon resulted in the emergence of a new subfield of machine learning that was described as relational learning~\cite{V05,R08}.
For instance, TILDE~\cite{BR98} learns decision trees within inductive logic programming systems. 
Serafino et al.~\cite{SPC18} proposed an ensemble learning-based relational learning model for multi-type classification tasks in HINs.
Petkovic et al.~\cite{PCKD20} proposed a relational feature ranking method based on the gradient-boosted relational trees towards relational data. 
Lavrac et al.~\cite{LSR20} presented a unifying methodology combining propositionalisation and embedding techniques, which benefit from the advantages of both in solving complex relational learning tasks. 
Nevertheless, most of them are not in virtue of neural networks, which fall behind in automatically mining complex HINs.

\smallskip\noindent
\textbf{Graph Neural Networks.}
Existing GNNs generalise the convolutional operations of deep neural networks to deal with arbitrary graph-structured data.
Generally, a GNN model can be regarded as using the input graph structure to generate the computation graph of nodes for message passing, the local neighbourhood information is aggregated to get more effective contextual node representations in the network~\cite{BZSL14,DBV16,KW17,VCCRLB18}.
However, the complex graph-structured data in the real world are commonly associated with multiple types of objects and relations.
All of the GNN models mentioned above assume homogeneous graphs, thus it is difficult to apply them to HINs directly.

\smallskip\noindent
\textbf{Heterogeneous Graph Representation Learning.}
HGRL aims to project nodes in a HIN into a low-dimensional vector space while preserving the heterogeneous node attributes and edges.
A recent survey presents a comprehensive overview of HGRL~\cite{DHWST20}, covering shallow heterogeneous network embedding methods~\cite{DCS17,FLL17,FHZYA18}, and heterogeneous GNN-based approaches that are empowered by rather complex deep encoders~\cite{MCMSZ15,SKBBTW18,ZSHSC19,WJSWYCY19,FZMK20,HDWS20,ZMGZ21}.
The ``shallow'' methods are characterised as an embedding lookup table, meaning that they directly encode each node in the network as a vector, and this embedding table is the parameter to be optimised.
However, they cannot utilise the node attributes and do not support the end-to-end training strategy.
On the other hand, inspired by the recent outstanding performance of GNN models, some studies have attempted to extend GNNs for HINs.
R-GCNs~\cite{SKBBTW18} keep a distinct linear projection weight for each relation type.
HetGNN~\cite{ZSHSC19} adopts different recurrent neural networks for different node types to incorporate multi-modal attributes.
HAN~\cite{WJSWYCY19} extends GAT~\cite{VCCRLB18} by maintaining weights for different meta-path-defined edges.
MAGNN~\cite{FZMK20} defines meta-path instance encoders, which are used to extract the structural and semantic information ingrained in the meta-path instances.
However, all of these models require manual effort and domain expertise to define meta-paths in order to capture the semantics underlying the given heterogeneous graph.

A recent model, HGT~\cite{HDWS20}, attempts to avoid the dependency on hand-crafted meta-paths by devising transferable relation scores, but the number of layers limits its exploration range, and it introduces a large number of additional parameters to optimise.
GTN~\cite{YJKKK19} selects meta-paths from a group of adjacency matrices with learnable weights.
The weights are shared among all nodes and are thus not flexible to generate node-specific meta-paths for each individual node.
In addition, FSPG~\cite{MCMSZ15}, AutoPath~\cite{YLHZPH18} and MPDRL~\cite{WDPH20} attempt to employ RL technologies to discover paths between pairs of nodes and further learn node representations for predicting the possibility of the existing edges between node pairs.
They assume the founded paths explicitly represent the similarity between two nodes.
However, they can only identify meta-paths that describe two nodes' similarities instead of generating meta-path for individual nodes to learn their representations for node-wise tasks.
Moreover, some work~\cite{TSRMW18,AHMOP20} concerns the discovery of frequent patterns in a HIN and the subsequent transformation of these patterns into rules, aka rule mining. 
But the found patterns are not designed for specific tasks or nodes. 
Consequently, we believe it is necessary and essential to developing a new HGRL framework that can support the adaptive generation of personalised meta-paths for each node in HIN for node-wise tasks. 

\smallskip\noindent
\textbf{Discussion}.
Table~\ref{table:comparison_diff_models} summarises the key advantages of PM-HGNN and compares it with a number of recent state-of-the-art methods.
PM-HGNN is the first HGRL model that can adaptively generate personalised meta-paths for each individual node to support node-wise tasks and maintain the end-to-end training mechanism. 

% section related_work (end)
%------------------------------------------------------------------------------

%------------------------------------------------------------------------------
\section{Preliminaries} % (fold)
\label{sec:preliminaries}
\subsection{Problem Statement}
\label{subsec:problem_statememt}

\textbf{Definition 1.} \hspace{.1mm} (\textbf{Heterogeneous Information Network}): A HIN is defined as a directed graph $G=(V, E, N, R)$, associated with a node type mapping function $\phi : V \rightarrow N$ and a relation type mapping function $\varphi: E \rightarrow R$, where $N$ and $R$ are the sets of node and edge types, respectively.
Node $v_{i}$'s attribute vector is denoted as $x_i \in \mathbb{R}^{\lambda}$ (with the dimensionality $\lambda$). 

\smallskip\noindent
\textbf{Definition 2.} \hspace{.1mm} (\textbf{Meta-path}):
Given a HIN $G$, a meta-path $\Omega$ with length $T$ is defined as: $\omega_{0} \xrightarrow{r_{1}} \omega_{1} \xrightarrow{r_{1}} \dots \xrightarrow{r_{T}} \omega_{T}$, where $\omega_{j} \in N$ denotes a certain node type, and $r_{i} \in R$ denotes a relation type. 

\smallskip\noindent
\textbf{Definition 3.} \hspace{.1mm} (\textbf{Meta-path Instance}):
Given a meta-path $\Omega$, a meta-path instance $p$ is defined as a node sequence following the schema defined by $\Omega$.

\smallskip\noindent
\textbf{Problem: Heterogeneous Graph Representation Learning.}
We formulate heterogeneous graph representation learning (HGRL) as an information integration optimisation problem.
For a given HIN $G$, $f: (V, E)\rightarrow \mathbb{R}^{d}$ be a function to integrate information from node attributes and network structure as node representations.
Without manually specifying any meta-paths, we aim to jointly generate $\rho$ (e.g., $\rho$=1) meta-paths $\{\Omega_{j}\}^{\rho}_{j=1}$ for each node $v_i\in V$ to guide $f$ to encode rich structural and semantic information in HIN, and accordingly learn representations for all nodes in $G$. 

\subsection{Markov Decision Process}
\label{subsec:intro_mdp}
Markov Decision Process (MDP) is an idealised mathematical form to describe a sequential decision process that satisfies the \emph{Markov Property}~\cite{SB98}.
An MDP can be formally represented with the quadruple $(S, A, \mathcal{P}, \mathcal{R})$, where $S$ is a finite set of states, $A$ is a finite set of actions, and $\mathcal{P}: S \times A \rightarrow (0, 1)$ is a decision policy function to identify the probability distribution of the next action according to the current state.
Specifically, the policy encodes the state and the available action at step $t$ to output a probability distribution $\mathcal{P}(a_{t} \vert s_{t})$, where $s_{t} \in S$ and $a_{t} \in A$.
$\mathcal{R}$ is a reward function $\mathcal{R}: S \times A \rightarrow \mathbb{R}$, evaluating the result of taking action $a_{t}$ on the observed state $s_{t}$.

\begin{figure}[ht!]
\centering
\includegraphics[width=.7\linewidth]{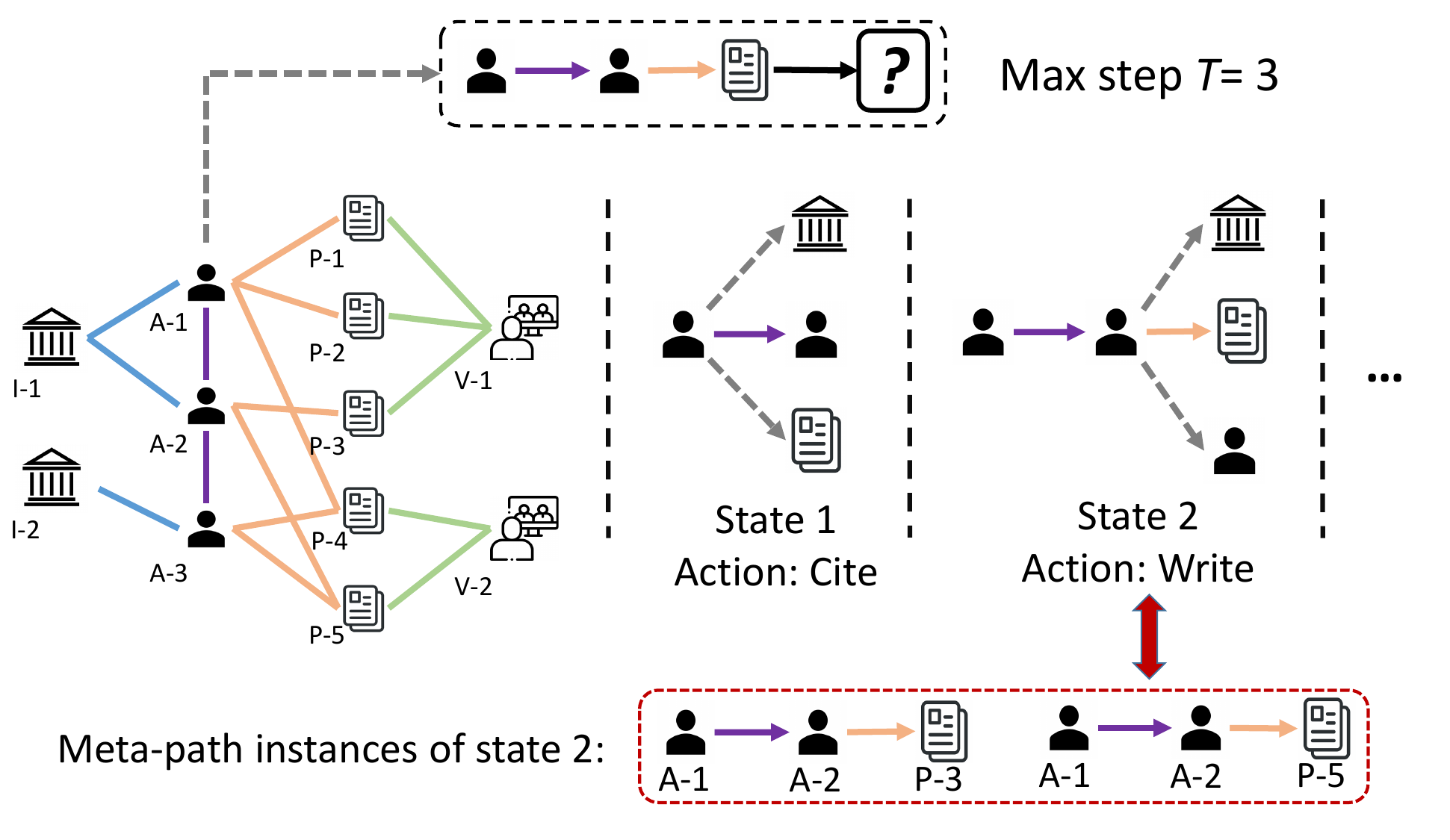}
\caption{
    Illustration of generating meta-paths as an MDP.
}
\label{fig:metapath_gene_mdp}
\vspace{-3mm}
\end{figure}

\smallskip\noindent
\textbf{Modelling HGRL with MDP}.
As illustrated in Fig.~\ref{fig:metapath_gene_mdp}, the meta-path generation process of HGRL can be naturally modelled as an MDP.
To generate a meta-path with maximum of 3 steps for $v_{\textit{A-1}}$, we take the first step as an example, the state $s_{1}$ is identifiable information of $v_{\textit{A-1}}$ and the action set includes relations in the HIN, i.e., $\{\it Work, Cite, \dots \}$.
The decision maker selects one relation from the action set to extend the meta-path as $a_{1} = {\it argmax}_{a \in A}(\mathcal{P}(a \mid s_{1}))$.
Then, the selected meta-path is fed into HGNN to learn node representation and apply it to the downstream task to obtain a reward score $\mathcal{R}(s_{1}, a_{1})$ that can be used to update $\mathcal{P}$.
We refer to Sec.~\ref{sec:methodology} for more details about modelling HGRL with MDP.

\subsection{Solving MDP with Reinforcement Learning}
\label{subsec:solve_mdp_with_dl}
Deep Reinforcement Learning (RL) is a family of algorithms that optimise the MDP with deep neural networks.
At each step $t$, the RL agent takes action $a_{t} \in A$ based on the current state $s_{t} \in S$, and observes the next state $s_{t+1}$ as well as a reward $r_{t} = \mathcal{R}(s_{t}, a_{t})$.
Looking at the definition of MDP, the agent acts as the decision policy with $\mathcal{P}$.
We aim to search for the optimal decisions that maximise the expected discounted cumulative reward, i.e., we aim to learn an agent network $\pi: S \rightarrow A $ to maximise $\mathbb{E}_{\pi} [\sum_{t'=t}^{T} \gamma^{t'}r_{t'}]$, where $T$ is the maximum number of steps, and $\gamma \in [0, 1]$ is a discount factor to balance short-term and long-term gains, and smaller $\gamma$ values place more emphasis on immediate rewards~\cite{ADBB17}.

Existing RL algorithms are mainly classified into two series: model-based and model-free algorithms.
Compared with model-based algorithms, model-free algorithms have better flexibility, as there is always the risk of suffering from model understanding errors, which in turn affects the learned policy~\cite{ADBB17}.
We adopt a classic model-free RL algorithm, i.e., Deep Q-learning (DQN)~\cite{MKSRVBGRF15}.
The basic idea of DQN is to estimate the action-value function by using the Bellman equation~\cite{MKSRVBGRF15} ($\mathcal{Q}^{*}$) as an interactive update based on the following intuition: if the optimal value $\mathcal{Q}^{*}(s_{t}, a_{t})$ of the state $s_{t}$  was known for all possible actions $a_{t} \in A$, then the optimal policy is to select the action $a_{i}$ maximising the expected value $\mathcal{R}(s_{t}, a_{t}) + \gamma \mathcal{Q}^{*}(s_{t+1}, a_{t+1})$.
And it is common to use a function approximator $\mathcal{Q}$ to estimate $\mathcal{Q}^{*}$~\cite{MKSRVBGRF15}:

\begin{equation}
\label{eq:qlearning}
    \mathcal{Q}(s_{t},a_{t}; \theta) = \mathbb{E}_{s_{t+1}} \left[\mathcal{R}(s_{t}, a_{t}) + \gamma \max_{a_{t+1}\in A} \left(\mathcal{Q}(s_{t+1}, a_{t+1}; \theta)\right)\right],
\end{equation}
where $\theta$ stands for the trainable parameters in the neural network that is used to estimate the decision policy.
A value iteration algorithm will approach the optimal action-value $\widetilde{\mathcal{Q}}$, i.e., $\mathcal{Q} \to \widetilde{\mathcal{Q}}$, as $t \to \infty$.

DQN exploits two techniques to stabilise the training:
(1) the \emph{memory buffer} $\mathcal{D}$ that stores the agent's experience in a replay memory, which can be accessed to perform the weight updating. 
(2) the \emph{separate target Q-network} ($\hat{\mathcal{Q}}$) to generate the target for Q-learning, which is periodically updated.

% section preliminaries (end)
%------------------------------------------------------------------------------

%------------------------------------------------------------------------------
\section{The PM-HGNN Framework} % (fold)
\begin{figure}[ht!]
\centering
\includegraphics[width=1.1\linewidth]{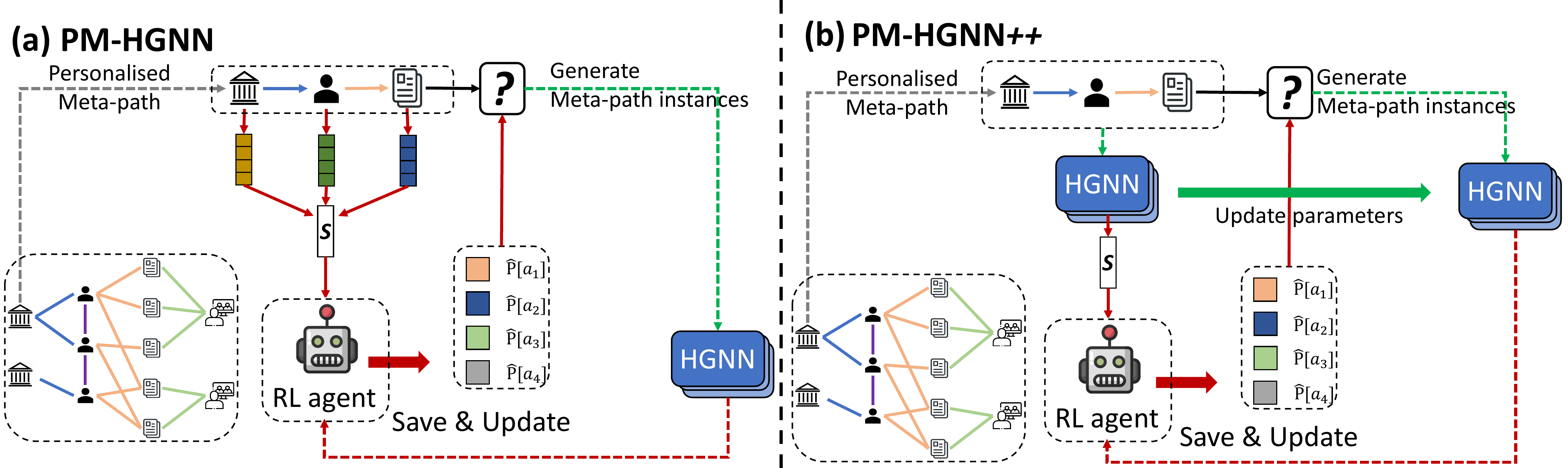}
\caption{
    Overview of PM-HGNN and PM-HGNN\textit{++}. 
}
\label{fig:general_structure}
\vspace{-3mm}
\end{figure}

\label{sec:methodology}
We present the overview of the proposed PM-HGNN in Fig.~\ref{fig:general_structure}-(a), which consists of two components: the RL agent module and an HGNN module.
According to states, the RL agent aims at predicting actions for each individual node to arrive at better rewards.
Next, we generate meta-path instances based on generated personalised meta-paths to support the information aggregation of HGNN to learn effective node representations. 
Finally, we apply generated representations on the downstream task for performance evaluation to obtain reward scores and save states, actions, and reward scores into the RL agent for the subsequent updating.

% In what follows, we elaborate on the details of our PM-HGNN framework.
% We first describe how we can generate meta-paths with the RL agent and train the RL agent.
% Then, we show how to perform information aggregation with the derived meta-paths for generating node representations.
% Last, we further present an extension model, PM-HGNN\textit{++}. 

\subsection{Personalised Meta-path Generation with RL}
\label{subsec:meta_path_generation_with_RL}
A personalised meta-path generation process with maximum $T$ steps for each node can be modelled as a $T$-round decision-making process that can be naturally treated as an MDP. 
We elaborate on the alignment between each MDP component and the personalised meta-path generation process in the following. 

\smallskip\noindent
\textbf{State ($S$)}: The state is a vector used to assist the decision policy to select a relation type to extend the personalised meta-paths for each node.
Hence, it is crucial to comprehensively encode existing parts of a meta-path into a state.
We adopt a gating mechanism to adaptively update the state. 
Take $v_{i}$'s meta-path $\Omega$ (starting from node type $N_{v_{i}}$) as an example, the state $s_{t}$ of $\Omega$ at step $t$ is formally defined as:
\begin{equation}
\label{eq:state_static}
    s_{t} = q \circ \left(\frac{1}{\vert D(v_{i}) \vert} \sum_{j \in D(v_i)} x_{j}\right) + (1-q) \circ s_{t-1},
\end{equation}
where $\circ$ stands for the Hadamard product, $D(v_{i})$ represents a set of past nodes at step $t$, and $\frac{1}{\vert D(v_{i}) \vert} \sum_{j \in D(v_{i})} x_{j}$ represents the average vector of past nodes' attributes.
$s_{t-1}$ stands for the state at step $t-1$.
$q$ is the update gate that can determine whether to update a state with past nodes' attributes, and we estimate it by exploring the relationship between past nodes' attributes and states.
It is formally defined as:
$q = Sigmoid\left(f_{\varphi}\left((\frac{1}{\vert D(v_i) \vert} \sum_{j \in D(v_i)} x_{j}) \parallel s_{t-1}\right)\right)$.
$f_{\varphi}$ can be seen as a shared convolutional kernel~\cite{VCCRLB18}, and $1-q$ is the reset gate.

\smallskip\noindent
\textbf{Action ($A$)}: 
The action space is a set of relation types to be chosen by the policy network to extend the meta-path, and each relation type is labelled with a positive integer.
Note that we add a special action \emph{STOP} to allow each node to have flexible-length meta-paths.
Beginning with the starting node $v_{i}$, the decision policy iteratively predicts which relation can lead to higher reward scores and accordingly use it to extend the current meta-path.
The decision policy selects the action \emph{STOP} to finish the path generation process if encountering a state that includes any extra relationship to the current meta-path hurts the performance on the downstream task.

\smallskip\noindent
\textbf{Decision Policy ($\mathcal{P}$)}:
The decision policy aims at mapping a state in $S$ into action in $A$.
The state space and the action space are continuous and discrete, respectively.
Thus we use a deep neural network to approximate the action-value function: $\mathcal{P}(a_{t} \vert s_{t}; \theta): S \times A \to (0, 1)$.
Besides, since any arbitrary action $a \in A$ is always a positive integer, we use DQN~\cite{MKSRVBGRF15} to optimise the meta-path generation problem.
We utilise an MLP~\cite{H99} as the decision policy network in the DQN, defined as:
% \begin{equation}
%     \begin{aligned}
%         z_{1} & = W^T_{1} s_{t} + c_{1}, \\
%         z_{2} & = W^T_{2} z_{1} + c_{2}, \\
%         & \dots \\
%         \hat{P} & = {\it Softmax}(\phi_{m}(W^T_{m} z_{m-1} + c_{m})),
%     \end{aligned}
% \end{equation}
$z_{1} = W^T_{1} s_{t} + b_{1}$, 
$z_{2} = W^T_{2} z_{1} + b_{2}$,
$...$,
$\hat{P} = {\it Softmax}(\phi_{m}(W^T_{m} z_{m-1} + b_{m}))$, 
where \(W_{m}\) and \(b_{m}\) denote the weight matrix and bias vector for the \(m\)-th layer's perceptron, respectively.
The output $\hat{P} \in (0, 1)$ stands for the possibilities of selecting different relations $a_{t} \in A$ to extend the meta-path. 
Note that it is possible to adopt other RL algorithms to optimise the policy network.
Here, we utilise a basic RL algorithm to illustrate our framework's main idea and demonstrate its effectiveness.

\smallskip\noindent
\textbf{Reward Function ($\mathcal{R}$)}: 
We devise a reward function to encourage the RL agent to achieve better and more stable performance on downstream tasks.
We define the reward function $\mathcal{R}$ as the performance improvement on the specific downstream task comparing with the historical performances, given by:
\begin{equation}
\label{eq:rfunc}
    \mathcal{R}(s_{t}, a_{t}) = \mathcal{M}(s_{t}, a_{t}) - \frac{\sum_{j=t-b}^{t-1} \mathcal{M}(s_{j}, a_{j})}{b},
\end{equation}
where $\frac{\sum_{i=t-b}^{t-1} \mathcal{M}(s_{j}, a_{j})}{b}$ is the baseline performance value at step $t$, which contains the historical performances of the last $b$ steps.
$\mathcal{M}(s_{t}, a_{t})$ is the performance based on the learned node representation $H^{t}[v_{i}]$ on the downstream task (e.g., node classification).
And use its accuracy on the validation set as the evaluation performance $\mathcal{M}$.

\smallskip\noindent
\textbf{Optimisation.}
The proposed meta-path generation 
at step $t$ consists of three phases: (i) obtaining state $s_{t}$; (ii) predicting an action $a_{t}=\arg\max_{a \in A}(Q(s_{t}, a; \theta))$ to extend the meta-path according to the current state $s_{t}$; (iii) updating state $s_{t}$ to $s_{t+1}$.
Moreover, we train the policy network Q-function by optimising Eq.~\ref{eq:qlearning} with the reward function as defined in Eq.~\ref{eq:rfunc}, and the loss function is given by: 
\begin{align}
\label{eq:loss_q}
    \mathcal{L}_{Q}(\theta) = \mathbb{E}_{\mathcal{T} \sim {U(\mathcal{D})}}[(\mathcal{R}(s_{t}, a_{t}) &+ \gamma \max_{a_{t+1} \in A} \hat{\mathcal{Q}}(s_{t+1}, a_{t+1}; \theta^{-}) \nonumber \\
    &\quad - \mathcal{Q}(s_{t}, a_{t}; \theta) )^{2}],
\end{align}
where $\mathcal{T} = (s_{t}, a_{t}, s_{t+1}, \mathcal{R}(s_t, a_t))$ is randomly sampled replay memory from the \emph{memory buffer} $\mathcal{D}$, $\theta^{-}$ is the set of parameters in the \emph{separate target Q-network} $\hat{Q}$, $\max\limits_{a_{t+1}\in A} \hat{Q}(s_{t+1}, a_{t+1}; \theta^{-})$ stands for the optimal target value, and $ Q(s_{t}, a_{t}; \theta)$ is the predicted reward value based on the training Q-network.
We optimise the Q-network's parameters by minimising the loss function.

\subsection{Information Aggregation with Personalised Meta-paths}
\label{subsec:info_agg}
Due to the heterogeneity of nodes in HIN, various types of nodes have different attribute spaces.
Before the information aggregation, we first do a node type-specific transformation to project the latent features of different node types into the same space, given by: $H^{0}[v_{i}] = W_{\omega_{i}} \cdot x_{i}$, where $x_{i}$ and $H^{0}[v_{i}]$ are the original attributes and projected features of node $v_{i}$, and $W_{\omega_{i}}$ is a learnable transformation matrix for the type node $\omega_{i} = \phi(v_i)$.
Next, we perform the information aggregation using the generated meta-path instances to learn effective node representations for the downstream task.
Take node $v_{i}$ as an example, and its corresponding obtained meta-path is assumed to be $\Omega$.

\begin{figure}[ht!]
\centering
\includegraphics[width=0.7\linewidth]{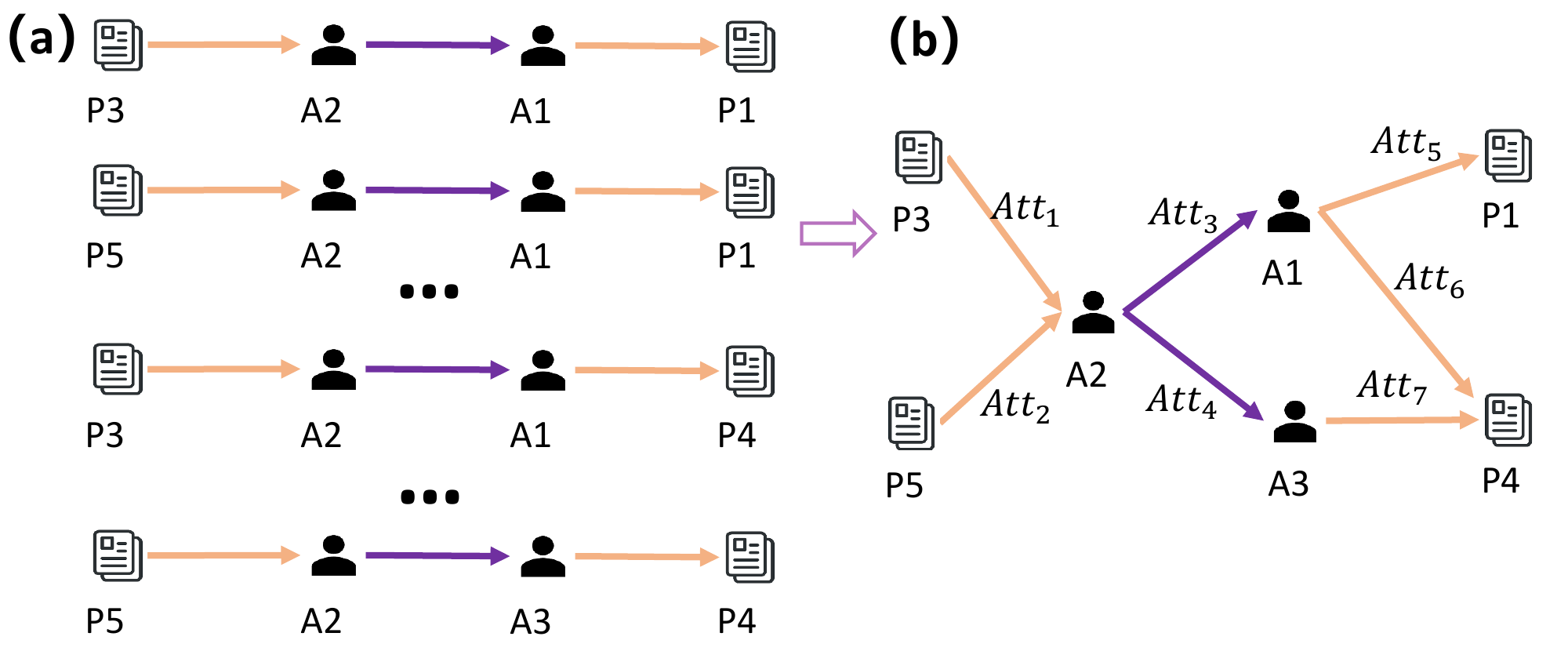}
\caption{
    Comparison between information aggregation by conventional meta-path instances and the proposed redundancy-free computation.
    (a) The sequential aggregation path instances generated from the meta-path 
    $\Omega$: {\it P$\xrightarrow{\it Written\_by}$A$\xrightarrow{\it Cite}$A$\xrightarrow{\it Write}$P} ({\it A}: {\it Author}, {\it P}: {\it Paper}).
    (b) The improved aggregation structure by our redundancy-free computation method with attention scores to distinguish messages from different nodes.
}
\label{fig:redundancy}
\vspace{-3mm}
\end{figure}

We first generate meta-path instances $\{p_{1}, p_{2}, \dots\}$ based on meta-path $\Omega$. 
As the example shown in Fig.~\ref{fig:redundancy}, there could be many meta-path instances that follow meta-path $\Omega$, (e.g., $\Omega: {\it Paper} \xrightarrow{\it Written\_by} {\it Author} \xrightarrow{\it Cite} {\it Author} \xrightarrow{\it Write} {\it Paper}$).
An intuitive approach to performing information aggregation is to adopt sequential aggregation, e.g., HetGNN~\cite{ZSHSC19} and HAN~\cite{WJSWYCY19}.
But we argue that these aggregated paths have computation redundancy, and the approach to aggregation can be further improved.
For example, $P_{3} \rightarrow A_{2}$ and $P_{5} \rightarrow A_{2}$ are repeatedly calculated in the first aggregation step. 
We aim to merge these two instances into one  process $\{P_{3}, P_{5} \} \rightarrow A_{2}$ to reduce redundant computations, termed as \emph{redundancy-free} aggregation. 
Besides, we learn the attention scores $Att(v_{i},v_{j})$, where nodes $v_{i}$ and $v_{j}$ are two ends of a link, for each link in the aggregation path so that messages on different nodes can be distinguished from each other.

Let $H^{0}[v_i]$ be the projected features of node $v_{i}$ involved in the instance set $\{p_{1}, p_{2}, \dots\}$ of meta-path $\Omega$.
We give the updating procedure of $H[v_{i}]$:
\begin{equation}
\label{eq:agg}
    H^{l}[v_{i}] = \mathop{\it Aggregate}\limits_{v_{j} \in I(v_{i})} \left({\it Att}(v_{i}, v_{j}) \cdot H^{l-1}[v_{j}]\right),
\end{equation}
where $I(v_{i})$ indicates the set of past nodes of $v_{i}$ in $\{p_{1}, p_{2}, \dots\}$, $l \in \{1, 2, \dots, t \}$ is the id of aggregator to perform information aggregation, and ${\it Aggregate}(\cdot) = Relu ({\it Mean}(\cdot))$. 
The operator ${\it Att(\cdot)}$ calculates the importance of the received messages using the relation between past messages and node features, given by: 
\begin{equation}
    {\it Att}(i, j) = \mathop{\it Softmax}\limits_{j \in I(v_{i})} \left(LeakyRelu \left(W H^{l-1}[v_{j}] \parallel W H^{l-1}[v_{i}]\right)\right),
\end{equation}
where $W$ are the trainable parameters, and $\parallel$ is the concatenation operator.
Sec.~\ref{subsec:model_analysis_study} presents an empirical study to reveal how the redundancy-free aggregation affects the time efficiency.

\begin{algorithm}[!ht]
\caption{
PM-HGNN
}
\label{alg:PM-HGNN}
\hspace*{\algorithmicindent} \textbf{Input:} HIN $G=(V, E)$, Max step $T$, Number of RL agent training epochs $K$, Number of HGNN training epochs $B$,  \\
\hspace*{\algorithmicindent} \textbf{Output:} Node representations $H^{T}$
\begin{algorithmic}[1]
\State Initialise memory buffer $\mathcal{D}$ and Q-function
\For{$k \gets \{1, 2, \dots, K \}$}
    Initialise information aggregators (HGNN)
    \For{$t \gets \{0,1,2,\dots,T \}$}
        \State Sample a batch of nodes $V_{j}$ from $V$
        \For{$v_{i} \in V_{j}$}
            \State Obtain state $s_{t}[v_i]$ via Eq.~\ref{eq:state_static};
            \State Get action $a_{t}[v_i] = {\it argmax}_{a \in A}\mathcal{Q}(s_{t}[v_i], a; \theta)$
            \State Extend $\Omega_{t}[v_i]$ with $a_{t}[v_i]$ and generate $\{p_0, p_1, \dots\}$
        \EndFor
        \For{$\beta \gets \{1, 2, \dots, B \}$}
            \State Perform information aggregation on $\{p_0, p_1, \dots\}$ via Eq.~\ref{eq:agg} 
            \State Optimise HGNN' parameters via Loss-function $\mathcal{L}$
        \EndFor
        \State Information aggregation on $p_{t}$ with trained HGNN via Eq.~\ref{eq:agg}
        \State Obtain the learned node representations $H^{t}$
        \State Obtain $\mathcal{R}(s_t, a_t)$ on validation dataset via Eq.~\ref{eq:rfunc}
        \State Store the quartuple $\mathcal{T}=(s_{t}, a_{t}, s_{t+1}, \mathcal{R}(s_t, a_t))$ into $\mathcal{D}$
        \State Optimise Q-function using the data in $\mathcal{D}$ via Loss-function-Q (Eq.~\ref{eq:loss_q})
    \EndFor
\EndFor
\end{algorithmic}
\end{algorithm}

\subsection{Model Training}
\label{subsec:model_training}

\textbf{Training of HGNN.}
Finally, the updated node representation $H^t[v_{i}]$ can be applied to downstream tasks.
Take semi-supervised node classification as an example, with a small fraction of labelled nodes, we can optimise the HGNN's parameters by minimising the Cross-Entropy loss $\mathcal{L}$ via back-propagation and gradient descent: $\mathcal{L} = - \sum\limits_{v \in \mathcal{V}_{L}} \sum\limits_{c=0}^{C-1}y_{v}[c] \cdot \log(H^{t}[v][c])$, where $\mathcal{V}_{L}$ stands for the set of nodes with labels, $C$ is the number of classes, $y_{v}$ is the one-hot label vector of node $v$ and $H^{t}[v]$ is the learned representation vector of node $v$.
Meanwhile, the learned node representation can be evaluated by task-specific evaluation matrix $\mathcal{M}_{\it eval}$ for obtaining a reward score. 
We summarise our PM-HGNN framework in Algorithm~\ref{alg:PM-HGNN}.
Note that PM-HGNN framework maintains the ability of HGNN in learning node representations for newly added nodes to the graph. 
The trained RL agent can adaptively generate a meta-path for the new node to compute its representations. 

\subsection{PM-HGNN\textit{++}}
\label{subsec:PM-HGNN_++}
As seen in previous sections, PM-HGNN adopts the RL agent to generate meta-paths for HGRL adaptively.
However, reviewing the overall workflow of PM-HGNN, we identify two limitations.
(1) PM-HGNN neglects the relational structures of HIN while generating meta-paths;
(2) HGNN requires a large number of epochs to train their parameters and obtain node representations, which restricts the efficiency of PM-HGNN.
To be more specific, PM-HGNN generates meta-paths by only considering the node attribute information from the HIN, i.e., state $s_{t}$ summarising the attributes of nodes involved in the meta-path as Eq.~\ref{eq:state_static}.
However, HIN's semantic structure is also important information to assist the meta-path generation.
In addition, similar to the training process of other deep neural networks~\cite{KW17}, the encoder of PM-HGNN needs a number of epochs to train their parameters to learn effective node representations in terms of a set of meta-path instances.
In this way, if HGNN needs $B$ epochs to train their parameters, then we need $T \times B$ epochs to complete a meta-path generation process with the maximum number of steps $T$.
Meanwhile, the RL agent also needs lots of explorations to obtain numerous samples ($\mathcal{T}=(s_{t}, a_{t}, s_{t+1}, \mathcal{R}(s_t, a_t))$) to train the policy network that can result in a combinatorial scale explosion.
In the end, following the average aggregation approach to defining node states, PM-HGNN is not able to distinguish node states with categorical attributes, as shown in Eq.~\ref{eq:state_static}. 

\smallskip\noindent
\textbf{New State ($S$)}.
To address the identified limitations, we propose an extension of PM-HGNN, PM-HGNN\textit{++}, which can utilise the structure information of HIN and significantly accelerate the meta-path generation process.
Our solution is to define a novel state instead of Eq.~\ref{eq:state_static}.
Because we notice that the encoder (i.e., HGNN) of PM-HGNN can summarise node attributes and topological structure information involved in meta-paths to assist meta-path generation.
In particular, we utilise the latest node representation vector $H^{t-1}[v_i]$ of meta-path's starting node $v_{i}$ as the state.
The new state ($S$) can be formally described as: 
% $s_{t} = {\it Normalise}(H^{t-1}[i])$, 
\begin{equation}
\label{eq:state_++}
    s_{t} = {\it Normalise}(H^{t-1}[i]),
\end{equation}
where ${\it Normalise}$ is a normaliser trained on the first $B$ generated states to convert the states into the same distribution.
Hence, input will be normalized as $\frac{H[v_{i}]-H_{mean}}{H_{std}}$, where $H_{mean}$ and $H_{std}$ is calculated by the first $B$ states.
Note that {\it normaliser} is a common trick used in deep RL for stabilising the training when the $S$ is very sparse~\cite{HLWXC17}.
In addition, $H^{t}[i]$ contains not only node attributes but also the relevant graph structure information of each node, hence $s_t$ of Eq.~\ref{eq:state_++} can distinguish different nodes with categorical attributes.
So, it endows PM-HGNN\textit{++} with the ability to handle graphs with diverse node attributes.

% \begin{algorithm}[ht!]
% \SetAlgoLined
% \KwIn{
% 	HIN $G=(V, E)$,
% 	Max timesteps $T$, Number of RL agent training epochs $K$,
% }
% \KwOut{
% 	Node representations $H^{T}$.
% }
% Initialise information aggregators (HGNN), memory buffer $\mathcal{D}$ and Q-function, \\ 
% \For{$k \gets \{0, 1, K-1\}$}{
%     % Initialise information aggregators, \\
%     Step $t$ = mode($k$, $T$) + 1, \\
%     Sample a batch of nodes $V_{j}$ from $V$, \\
%     \For{$v_{i} \in V_{j}$}{
%         % Obtain state $s_{t}[v_i]$; \\
%         Obtain state $s_{t}[v_i]$ via Eq.~\ref{eq:state_++}; \\
%         Get the action $a_{t}[v_i] = argmax_{a \in A}Q(s_{t}[v_i], a; \theta)$; \\
%         Extend the meta-path $\Omega_{t}[v_i]$ with $a_{t}[v_i]$ and generate meta-path instances $\{p_0, p_1, \dots\}$. \\
%     }
%     Perform information aggregation on $\{p_0, p_1, \dots\}$ via Eq.~\ref{eq:agg}; \\
%     Obtain the learned node representations $H^{t}$; \\
%     Obtain $\mathcal{R}(s_t, a_t)$ on validation dataset via Eq.~\ref{eq:rfunc}; \\
%     Store the quartuple $\mathcal{T}=(s_{t}, a_{t}, s_{t+1}, R_{t})$ into $\mathcal{D}$; \\
%     Optimise HGNN's parameters via Loss-function $\mathcal{L}$; \\ 
%     % (Eq.~\ref{eq:loss_f}); \\
%     Optimise Q-function using the data in $\mathcal{D}$ via Loss-function-Q (Eq.~\ref{eq:loss_q}).
% }
% \caption{PM-HGNN\textit{++}}
% \label{alg:PM-HGNN-++}
% % \vspace{-3mm}
% \end{algorithm}
\begin{algorithm}[!ht]
\caption{
PM-HGNN\textit{++}
}
\label{alg:PM-HGNN-++}
\hspace*{\algorithmicindent} \textbf{Input:} HIN $G=(V, E)$, Max timesteps $T$, Number of RL agent training epochs $K$,  \\
\hspace*{\algorithmicindent} \textbf{Output:} Node representations $H^{T}$
\begin{algorithmic}[1]
\State Initialise information aggregators (HGNN), memory buffer $\mathcal{D}$ and Q-function
\For{$k \gets \{0, 1, K-1\}$}
    \State Step $t$ = mode($k$, $T$) + 1
    \State Sample a batch of nodes $V_{j}$ from $V$
    \For{$v_{i} \in V_{j}$}
        \State Obtain state $s_{t}[v_i]$ via Eq.~\ref{eq:state_++}
        \State Get the action $a_{t}[v_i] = argmax_{a \in A}Q(s_{t}[v_i], a; \theta)$
        \State Extend the meta-path $\Omega_{t}[v_i]$ with $a_{t}[v_i]$
        \State Generate meta-path instances $\{p_0, p_1, \dots\}$
    \EndFor
    \State Perform information aggregation on $\{p_0, p_1, \dots\}$ via Eq.~\ref{eq:agg}
    \State Obtain the learned node representations $H^{t}$
    \State Obtain $\mathcal{R}(s_t, a_t)$ on validation dataset via Eq.~\ref{eq:rfunc}
    \State Store the quartuple $\mathcal{T}=(s_{t}, a_{t}, s_{t+1}, R_{t})$ into $\mathcal{D}$
    \State Optimise HGNN's parameters via Loss-function $\mathcal{L}$
    \State Optimise Q-function using the data in $\mathcal{D}$ via Loss-function-Q (Eq.~\ref{eq:loss_q})
\EndFor
\end{algorithmic}
\end{algorithm}

\smallskip\noindent
\textbf{New Training Process}.
The training process can be further improved based on the new definition of the state.
The learned node representation can be further used to optimise HGNN's parameters and update the RL agent.
We achieve such a kind of mutual optimisation between the RL agent and HGNN since the meta-paths are generated based on the current status of HGNN.
With this novel training process, we only need $T$ epochs to complete a meta-path generation procedure because we do not need to wait for the encoder to complete an entire training process, instead of $T \times B$ epochs.
We summarise PM-HGNN\textit{++} in Algorithm~\ref{alg:PM-HGNN-++}. 
We perform empirical analysis to compare the time consuming of PM-HGNN and PM-HGNN\textit{++} in Sec.~\ref{subsec:model_analysis_study}.
Note that, similar to PM-HGNN, PM-HGNN\textit{++} framework can adaptively generate a meta-path for the new node to compute its representation, which maintains the ability of HGNN to learn representations for newly added nodes. 

\subsection{Model Analysis}
\label{subsec:model_analysis}
The proposed models can deal with various types of nodes and relations and fuse rich semantics in HINs. 
Benefiting from the RL agent, personalised meta-paths are generated for different nodes, and the HGNN encoder allows information to transfer between nodes via diverse relations. 
The RL agent that we adopted in this paper, i.e., Deep Q-learning (DQN), is hard to give the precise computation complexity~\cite{FWXY20}, hence we give the empirical meta-path generation time in IMDb and DBLP datasets in Sec.~\ref{subsec:model_analysis_study} (Fig.~\ref{fig:run_time_analysis}).
Given a meta-path $\Omega$ and the dimensionality of (hidden) node representations is $d$. 
The time complexity of the node representation learning process is $\mathcal{O}(V_{\Omega} d^2 + E_{\Omega}d)$, where $V_{\Omega}$ is the number of nodes following the meta-path $\Omega$ and $E_{\Omega}$ is the number of meta-paths based node pairs. 
$\mathcal{O}(V_{\Omega} d^2)$ accounts for the representation transformation process and $\mathcal{O}(E_{\Omega}d)$ represents the $Att(\cdot)$ computation process of relevant edges. 
To further unfold the relationship between the complexity of generated meta-paths and the performance, we also report how the maximum step $T$ and the number of aggregation paths affect the node classification performance based on the IMDb dataset in Sec.~\ref{subsec:model_analysis_study} (Table~\ref{fig:max_timestep_analysis}). 

% section methodology (end)
%------------------------------------------------------------------------------

%------------------------------------------------------------------------------
\section{Experiments} % (fold)
\label{sec:experiments}
\subsection{Experimental Settings}
\label{subsec:experimental_settings}

\textbf{Datasets.}
We adopt three HIN datasets (IMDb and DBLP, ACM)~\cite{WJSWYCY19,FZMK20} from different domains to evaluate our models' performance.
Statistic information is summarised in Table~\ref{table:dataset} and the meta-relation schema are shown in Fig.~\ref{fig:meta-relation_schema}.
The detailed description of datasets refers to Appendix~\ref{sec:appendix_datasets}.

\smallskip\noindent
\textbf{Competing Methods and Model Configuration.}
% To validate the effectiveness of the proposed models, w
We compare their performance against various state-of-the-art models.
They include $5$ homogeneous graph representation learning models: LINE~\cite{TQWZYM15}, DeepWalk~\cite{PAS14}, MLP, GCN~\cite{KW17}, GAT~\cite{VCCRLB18}; and $10$ HGRL models: Esim~\cite{SQLKHP16}, metapath2vec~\cite{DCS17}, JUST~\cite{HYC18}, HERec~\cite{SHZY19}, NSHE~\cite{ZWSLY20}, RGCN~\cite{SKBBTW18}, HAN~\cite{WJSWYCY19}, GTN~\cite{YJKKK19}, MAGNN~\cite{FZMK20} and HGT~\cite{HDWS20};
and $1$ state-of-the-art relational learning model PropStar~\cite{LSR20}. 
Note that FSPG~\cite{MCMSZ15} AutoPath~\cite{YLHZPH18} and MPDRL~\cite{WDPH20} do not support node-wise tasks since they learn paths that explicitly represent the similarity between pairs of nodes (as discussed in Sec.~\ref{sec:related_work}).
Therefore we cannot compare them.
The detailed model description and configuration for implementation refer to Appendix~\ref{sec:appendix_competing_methods} and Appendix~\ref{sec:appendix_implementation_details}.

\smallskip\noindent
\textbf{Evaluation Settings.}
We evaluate our models under node classification.
For the semi-supervised setting, we adopt the same settings as in MAGNN~\cite{FZMK20} to use $400$ instances for training and another $400$ instances for validation, the rest nodes for testing.
The generated node representations are further fed into a \emph{support vector machine} (SVM) classifier to get the prediction results.
For the supervised setting, we use the same percentage of nodes for training and validation, respectively. Use the rest of the nodes for testing.
% We employ an end-to-end mechanism, namely the model is trained and tuned on the training and validation sets, respectively, and generates the predictions on the testing set. 
We report the average \emph{Micro-F1} and \emph{Macro-F1} scores of $10$ runs with different seeds.

\subsection{Experimental Results}
\label{subsec:experimental_results}

\begin{table}[ht!]
\caption{
    Experiment results (\%) on the IMDb, DBLP and ACM datasets for the node classification task with unsupervised and semi-supervised settings.
    MP2V stands for metapath2vec.
    Best performance is marked in \textbf{bold}. 
}
\label{table:res_node_class}
\resizebox{1.1\textwidth}{!}{
\begin{tabular}{l c |ccccccc|ccccccccc}
% \toprule
\hline
                        & & \multicolumn{7}{|c|}{Unsupervised (w/ SVM)} & \multicolumn{8}{c}{Semi-supervised}   \\ 
                        % \cmidrule{3-17}
                        \hline
Dataset & Train & LINE & DeepWalk & ESim & MP2V & JUST & HERec & NSHE & GCN & GAT & HAN & GTN & HGT & PropStar & MAGNN & Ours & Ours\textit{++}          \\ 
                        % \midrule
                        \hline
\multirow{4}{*}{\specialcell{IMDb \\ Micro-F1}}
                        & 20\% & 45.21 & 49.94 & 49.32 & 47.22 & 50.62 & 46.23 & 51.69 & 52.80 & 53.64 & 56.32 & 58.64 & 57.49 & 58.23 & 59.60 & 62.14 & \textbf{62.53} \\ 
                        % \cmidrule{3-17}
                        % \hline
                        & 40\% & 46.92 & 51.77 & 51.21 & 48.17 & 52.25 & 47.89 & 53.17 & 53.76 & 55.56 & 57.32 & 59.73 & 58.27 & 59.29 & 60.50 & 62.18 & \textbf{63.10} \\ 
                        % \cmidrule{3-17}
                        % \hline
                        & 60\% & 48.35 & 52.79 & 52.53 & 49.87 & 52.28 & 48.19 & 54.04 & 54.23 & 56.47 & 58.42 & 60.64 & 59.22 & 59.80 & 60.88 & 62.69 & \textbf{63.13} \\ 
                        % \cmidrule{3-17}
                        % \hline
                        & 80\% & 48.98 & 52.72 & 52.54 & 50.50 & 53.54 & 49.11 & 54.25 & 54.63 & 57.40 & 59.24 & 61.58 & 60.16 & 60.24 & 61.53 & 62.86 & \textbf{64.07} \\ 
                        % \cmidrule{2-18}
                        \cline{2-18}
\multirow{4}{*}{\specialcell{IMDb \\ Macro-F1}}
                        & 20\% & 44.04 & 49.00 & 48.37 & 46.05 & 50.33 & 45.61 & 51.93 & 52.73 & 53.64 & 56.19 & 58.39 & 57.32 & 58.03 & 59.35 & 61.87 & \textbf{62.37} \\ 
                        % \cmidrule{3-17}
                        & 40\% & 45.45 & 50.63 & 50.09 & 47.57 & 52.04 & 46.80 & 53.75 & 53.67 & 55.50 & 56.15 & 59.82 & 58.19 & 58.06 & 60.27 & 62.17 & \textbf{62.95} \\ 
                        % \cmidrule{3-17}
                        & 60\% & 47.09 & 51.65 & 51.45 & 48.17 & 53.22 & 46.84 & 54.52 & 54.24 & 56.46 & 57.29 & 60.35 & 59.31 & 59.31 & 60.66 & 62.50 & \textbf{62.98} \\ 
                        % \cmidrule{3-17}
                        & 80\% & 47.49 & 51.49 & 51.37 & 49.99 & 53.67 & 47.73 & 54.38 & 54.77 & 57.43 & 58.51 & 61.17 & 60.14 & 59.93 & 61.44 & 62.65 & \textbf{63.98} \\ 
                        % \midrule
                        \hline
\multirow{4}{*}{\specialcell{DBLP \\ Micro-F1}}
                        & 20\% & 87.68 & 87.21 & 91.21 & 89.02 & 90.40 & 91.49 & 92.20 & 88.51 & 91.61 & 92.33 & 93.41 & 92.60 & 91.79 & 93.61 & 91.73 & \textbf{94.34}\\ 
                        % \cmidrule{3-17}
                        & 40\% & 89.25 & 88.51 & 92.05 & 90.36 & 91.27 & 92.05 & 92.63 & 89.22 & 91.77 & 92.57 & 93.50 & 93.01 & 92.04 & 93.68 & 92.01 & \textbf{95.78}\\ 
                        % \cmidrule{3-17}
                        & 60\% & 89.34 & 89.09 & 92.28 & 90.94 & 91.82 & 92.66 & 92.77 & 89.57 & 91.97 & 92.72 & 93.68 & 93.20 & 92.49 & 93.99 & 92.67 & \textbf{95.92}\\ 
                        % \cmidrule{3-17}
                        & 80\% & 89.96 & 89.37 & 92.68 & 91.31 & 92.15 & 92.78 & 93.20 & 90.33 & 92.24 & 93.23 & 94.32 & 93.97 & 93.16 & 94.47 & 93.91 & \textbf{96.63} \\ 
                        % \cmidrule{2-18}
                        \cline{2-18}
\multirow{4}{*}{\specialcell{DBLP \\ Macro-F1}}
                        & 20\% & 87.16 & 86.70 & 90.68 & 88.47 & 90.18 & 90.82 & 91.89 & 88.00 & 91.05 & 91.69 & 92.97 & 91.74 & 91.35 & 93.13 & 91.63 & \textbf{94.94}\\ 
                        % \cmidrule{3-17}
                        & 40\% & 88.85 & 88.07 & 91.61 & 89.91 & 90.44 & 91.44 & 91.82 & 89.00 & 91.24 & 91.96 & 92.88 & 92.21 & 92.07 & 93.23 & 92.06 & \textbf{95.42}\\ 
                        % \cmidrule{3-17}
                        & 60\% & 88.93 & 88.69 & 91.84 & 90.50 & 91.10 & 92.08 & 92.15 & 89.43 & 91.42 & 92.14 & 93.30 & 92.81 & 92.41 & 93.57 & 92.76 & \textbf{95.57}\\ 
                        % \cmidrule{3-17}
                        & 80\% & 89.51 & 88.93 & 92.27 & 90.86 & 91.96 & 92.25 & 92.39 & 89.98 & 91.73 & 92.50 & 94.02 & 93.16 & 92.82 & 94.10 & 94.89 & \textbf{96.30}\\ 
                        % \midrule
                        \hline
\multirow{4}{*}{\specialcell{ACM \\ Micro-F1}}
                        & 20\% & 84.41 & 84.22 & 87.10 & 85.00 & 87.76 & 88.13 & 87.10 & 86.12 & 88.45 & 88.89 & 89.06 & 88.67 & 87.82 & 90.16 & 89.62 & \textbf{91.38}\\ 
                        % \cmidrule{3-17}
                        & 40\% & 85.14 & 86.33 & 88.53 & 87.36 & 88.36 & 89.28 & 88.53 & 86.64 & 88.68 & 89.06 & 89.71 & 89.03 & 88.36 & 90.53 & 90.51 & \textbf{91.53}\\ 
                        % \cmidrule{3-17}
                        & 60\% & 86.03 & 86.95 & 89.64 & 88.75 & 89.85 & 89.90 & 89.08 & 87.00 & 88.81 & 89.92 & 90.95 & 90.09 & 88.65 & 90.87 & 90.90 & \textbf{91.98}\\ 
                        % \cmidrule{3-17}
                        & 80\% & 87.16 & 87.12 & 89.75 & 88.97 & 90.05 & 90.09 & 89.74 & 88.04 & 89.41 & 90.04 & 91.60 & 90.62 & 89.21 & 91.01 & 91.01 & \textbf{92.99}\\ 
                        % \cmidrule{2-18}
                        \cline{2-18}
\multirow{4}{*}{\specialcell{ACM \\ Macro-F1}}
                        & 20\% & 83.41 & 83.13 & 86.12 & 84.36 & 86.49 & 87.58 & 86.62 & 85.61 & 87.16 & 87.93 & 88.12 & 87.20 & 87.64 & 89.06 & 88.78 & \textbf{90.03}\\ 
                        % \cmidrule{3-17}
                        & 40\% & 84.06 & 85.37 & 87.82 & 86.63 & 87.15 & 88.28 & 87.12 & 85.93 & 87.43 & 88.05 & 88.67 & 88.14 & 87.95 & 89.38 & 89.46 & \textbf{91.08}\\ 
                        % \cmidrule{3-17}
                        & 60\% & 85.98 & 85.73 & 88.08 & 87.48 & 88.39 & 88.95 & 88.85 & 86.66 & 87.71 & 88.26 & 90.85 & 89.09 & 87.48 & 89.49 & 89.61 & \textbf{91.12}\\ 
                        % \cmidrule{3-17}
                        & 80\% & 86.88 & 85.85 & 89.64 & 88.37 & 88.70 & 89.06 & 89.50 & 87.00 & 88.02 & 88.88 & 90.90 & 90.02 & 88.59 & 90.71 & 90.25 & \textbf{92.62}\\ 
                        % \botrule
                        \hline
\end{tabular}
}
\vspace{-3mm}
\end{table}

\noindent
\textbf{Semi-supervised Node Classification}.
We present the results in Table~\ref{table:res_node_class}, in which competing methods that cannot support semi-supervised settings utilise the unsupervised settings.
PM-HGNN\textit{++} performs consistently better than all competing methods across different proportions and datasets.
On IMDb, the performance gain obtained by PM-HGNN\textit{++} over the best competing method (MAGNN) is around ($3.7\%-5.08\%$).
GNN models designed for homogeneous and heterogeneous graphs perform better than shallow HGRL models.
This demonstrates that the modelling of heterogeneous node attributes improves the performance significantly.
On DBLP and ACM, the performances of all models are overall better compared with the results on IMDb.
It is interesting to observe that different from the results on IMDb, the shallow heterogeneous network methods, i.e., JUST and NSHE obtain better performances than a few number of homogeneous GNNs.
That said, the heterogeneous relations on DBLP and ACM are useful for the node classification task.
PM-HGNN\textit{++} apparently outperforms the strongest competing method (i.e., MAGNN) up to $2.4\%$, showing the generality and superiority of PM-HGNN\textit{++}. 
In addition, among unsupervised competing methods, NSHE obtains better performance than other unsupervised methods, showing that network schema in HIN is helpful to obtain better representations.

\begin{table}[ht!]
\caption{
    Experiment results (\%) on the IMDb, DBLP and ACM datasets for the node classification task with supervised settings.
    Best performance is marked in \textbf{bold}. 
}
% \vspace{-3mm}
\label{table:res_node_class_sup}
% \footnotesize
% \scriptsize
% \small
\centering
\resizebox{1.1\linewidth}{!}{
\begin{tabular}{cc|ccccccccc|cc}
% \toprule
\hline
Dataset & Train & MLP & GCN & GAT & RGCN & HAN & GTN & HGT & PropStar & MAGNN & Ours & Ours\textit{++}          \\ 
                        % \midrule
                        \hline
\multirow{4}{*}{\specialcell{IMDb \\ Micro-F1}}
                        & 5\%  & 47.60 & 52.18 & 53.32 & 54.09 & 55.74 & 55.42 & 55.80 & 55.28 & 54.77 & 54.90 & \textbf{57.09} \\ 
                        % \cmidrule{3-12}
                        & 10\% & 49.75 & 55.81 & 56.53 & 58.22 & 59.21 & 60.11 & 60.06 & 59.82 & 59.26 & 60.97 & \textbf{62.27} \\ 
                        % \cmidrule{3-12}
                        & 15\% & 54.62 & 58.55 & 59.13 & 60.13 & 60.25 & 60.76 & 60.67 & 60.83 & 60.03 & 61.12 & \textbf{63.77} \\ 
                        % \cmidrule{3-12}
                        & 20\% & 55.22 & 59.13 & 60.22 & 60.15 & 60.70 & 61.58 & 61.71 & 61.77 & 60.21 & 62.33 & \textbf{64.65} \\ 
                        % \cmidrule{2-13}
                        \cline{2-13}
\multirow{4}{*}{\specialcell{IMDb \\ Macro-F1}}
                        & 5\%  & 47.67 & 51.89 & 53.17 & 54.20 & 55.53 & 58.39 & 55.36 & 55.16 & 54.36 & 54.31 & \textbf{56.47} \\ 
                        % \cmidrule{3-12}
                        & 10\% & 49.07 & 54.96 & 56.32 & 58.17 & 58.57 & 59.82 & 60.13 & 60.19 & 58.22 & 61.02 & \textbf{62.04} \\ 
                        % \cmidrule{3-12}
                        & 15\% & 54.86 & 57.23 & 58.82 & 60.02 & 60.03 & 60.35 & 60.42 & 60.60 & 59.76 & 60.93 & \textbf{63.81} \\ 
                        % \cmidrule{3-12}
                        & 20\% & 55.02 & 58.56 & 59.90 & 60.14 & 60.44 & 61.17 & 61.43 & 61.86 & 60.28 & 62.07 & \textbf{63.89} \\
                        % \midrule
                        \hline
\multirow{4}{*}{\specialcell{DBLP \\ Micro-F1}}
                        & 5\%  & 69.25 & 78.94 & 78.01 & 81.98 & 90.82 & 90.90 & 90.08 & 90.80 & 90.22 & 89.11 & \textbf{92.52} \\ 
                        % \cmidrule{3-12}
                        & 10\% & 75.75 & 82.53 & 82.69 & 82.73 & 92.17 & 92.25 & 92.20 & 92.57 & 91.96 & 90.43 & \textbf{93.87} \\ 
                        % \cmidrule{3-12}
                        & 15\% & 76.87 & 83.50 & 84.61 & 83.29 & 92.62 & 92.77 & 92.82 & 92.93 & 92.31 & 91.20 & \textbf{94.72} \\ 
                        % \cmidrule{3-12}
                        & 20\% & 77.95 & 83.94 & 85.07 & 84.67 & 93.16 & 93.24 & 93.21 & 93.45 & 93.02 & 91.98 & \textbf{95.13} \\ 
                        % \cmidrule{2-13}
                        \cline{2-13}
\multirow{4}{*}{\specialcell{DBLP \\ Macro-F1}}
                        & 5\%  & 68.50 & 78.23 & 76.74 & 81.11 & 90.24 & 90.32 & 89.93 & 89.98 & 90.06 & 88.63 & \textbf{92.20} \\ 
                        % \cmidrule{3-12}
                        & 10\% & 74.85 & 81.46 & 81.59 & 82.65 & 91.59 & 91.99 & 91.21 & 91.12 & 91.31 & 89.70 & \textbf{93.34} \\ 
                        % \cmidrule{3-12}
                        & 15\% & 76.17 & 82.77 & 82.00 & 83.79 & 91.90 & 92.45 & 92.90 & 92.30 & 92.09 & 90.84 & \textbf{94.14} \\ 
                        % \cmidrule{3-12}
                        & 20\% & 76.86 & 83.05 & 84.34 & 84.11 & 92.13 & 93.03 & 92.97 & 93.05 & 92.57 & 91.31 & \textbf{94.33} \\ 
                        % \midrule
                        \hline
\multirow{4}{*}{\specialcell{ACM \\ Micro-F1}} 
                        & 5\%  & 70.30 & 76.23 & 77.35 & 79.35 & 88.83 & 88.27 & 88.50 & 88.35 & 88.29 & 88.86 & \textbf{90.53} \\ 
                        % \cmidrule{3-12}
                        & 10\% & 73.55 & 80.85 & 80.28 & 80.95 & 90.08 & 90.18 & 90.16 & 89.52 & 90.37 & 89.31 & \textbf{92.72} \\ 
                        % \cmidrule{3-12}
                        & 15\% & 75.22 & 82.33 & 82.26 & 82.46 & 91.84 & 91.66 & 91.87 & 90.11 & 90.76 & 89.77 & \textbf{92.88} \\ 
                        % \cmidrule{3-12}
                        & 20\% & 76.18 & 82.96 & 84.32 & 83.08 & 92.56 & 92.06 & 91.96 & 91.56 & 91.22 & 90.99 & \textbf{93.93} \\ 
                        % \cmidrule{2-13}
                        \cline{2-13}
\multirow{4}{*}{\specialcell{ACM \\ Macro-F1}} 
                        & 5\%  & 71.80 & 77.01 & 77.53 & 80.85 & 88.73 & 87.57 & 89.51 & 89.79 & 89.46 & 88.62 & \textbf{90.47} \\ 
                        % \cmidrule{3-12}
                        & 10\% & 72.56 & 79.23 & 79.60 & 81.37 & 89.40 & 88.99 & 90.12 & 90.15 & 90.02 & 89.06 & \textbf{91.20} \\ 
                        % \cmidrule{3-12}
                        & 15\% & 73.73 & 80.42 & 80.39 & 81.79 & 90.73 & 89.58 & 90.80 & 90.87 & 90.42 & 89.21 & \textbf{92.04} \\ 
                        % \cmidrule{3-12}
                        & 20\% & 74.86 & 81.07 & 81.64 & 82.86 & 91.80 & 90.71 & 91.40 & 91.13 & 91.32 & 90.87 & \textbf{93.55} \\ 
                        % \botrule
                        \hline
\end{tabular}
}
\vspace{-3mm}
\end{table}

\smallskip\noindent
\textbf{Supervised Node Classification}.
We present the results in Table~\ref{table:res_node_class_sup}.
We can see that PM-HGNN\textit{++} consistently outperforms all competing models on IMDb, DBLP and ACM datasets, with up to $5.6\%$ in terms of \emph{Micro-F1}.
Heterogeneous GNNs outperform homogeneous GNNs, and our PM-HGNN\textit{++} achieves the best performance.
This demonstrates the usefulness of heterogeneous relations and the advantages of generating appropriate personalised meta-paths for each node according to its attributes and relational structure.

\subsection{Meta-path Analysis}
\label{subsec:meta-paths_analysis}

\noindent
\textbf{Meta-path Generation Process Visualisation.}
We visualise how the RL agent in PM-HGNN\textit{++} generates personalised meta-paths for each target node in Fig.~\ref{fig:action_analysis_IMDb}, Fig.~\ref{fig:action_analysis_DBLP} and Fig.~\ref{fig:action_analysis_ACM}.
Fig.~\ref{fig:action_analysis_IMDb} summarises the actions made by the RL agent on IMDb with different max steps under the semi-supervised setting.
The percentages marked in the figure present the fraction of nodes choosing the corresponding relation to extend the meta-path at that step.
For example, the meta-path ${\it M\xrightarrow{64.8\%}A\xrightarrow{8.8\%}M}$ with $T=2$.
$64.8\%$ means there are $64.8\%$ {\it Movie} nodes select {\it MA} relation at step-$1$, and $8.8\%$ means among those {\it Movie} nodes who selected {\it MA} at step-$1$, there are $8.8\%$ nodes selecting {\it AM} to extend the meta-path.
It is interesting to see that the RL agent selects the relation {\it MA} at the first step more often than the other one {\it MD}. 
It means that a movie's characteristic is more related to its actors than its director.
Besides, when step $T=2$, the RL agent selects plenty of \emph{STOP} actions.
This shows that two short meta-paths, i.e., ${\it Movie}\xrightarrow{\it MA} {\it Actor}$ and ${\it Movie} \xrightarrow{\it MD} {\it Director}$ are informative enough for the majority of nodes to learn effective representations.
Moreover, there are $6.7\%-9.4\%$ nodes that do not need any meta-paths to learn their representations.
This implies that their attributes can provide enough information.
This has also been reflected in the results of MLP in Sec.~\ref{subsec:experimental_results}, which does not utilise any structural information but only node attributes. 
More analyses on Fig.~\ref{fig:action_analysis_DBLP} and Fig.~\ref{fig:action_analysis_ACM} can be found in Appendix~\ref{sec:appendix_more_meta-paths_analysis}. 

\begin{table}[!ht]
\caption{Meta-paths designed by PM-HGNN\textit{++} on IMDb.}
% \vspace{-3mm}
\label{table:meta_paths_IMDb}
% \scriptsize
\centering
% \resizebox{.65\linewidth}{!}{
\begin{tabular}{c|c|c}
    % \toprule
    \hline
    $T$                 & \specialcell{Meta-paths \\ Designed by PM-HGNN\textit{++}} & Percentage (\%) \\ 
    % \midrule
    \hline
    $T$ = 1             & \specialcell{ {\it Movie $\rightarrow$ Actor} \\ {\it Movie $\rightarrow$ Director} } & \specialcell{ 58.1 \\ 32.9 } \\ 
    % \midrule
    \hline
    $T$ = 2             & \specialcell{ {\it Movie $\rightarrow$ Actor} \\ {\it Movie$\rightarrow$ Director} \\ {\it Movie$\rightarrow$Actor$\rightarrow$Movie} \\ {\it Movie$\rightarrow$Director$\rightarrow$Movie} } & \specialcell{59.1 \\ 23.8 \\ 5.7 \\ 4.7} \\ 
    % \midrule
    \hline
    $T$ = 3             & \specialcell{{\it Movie$\rightarrow$Actor} \\ {\it Movie$\rightarrow$Director} \\ {\it Movie$\rightarrow$Actor$\rightarrow$Movie$\rightarrow$Actor} \\ {\it Movie$\rightarrow$Actor$\rightarrow$Movie$\rightarrow$Director} \\ {\it Movie$\rightarrow$Actor$\rightarrow$Movie}} & \specialcell{42.7 \\ 25.8 \\ 5.9 \\ 4.9 \\ 3.2} \\ 
    % \toprule
    \hline
    \specialcell{Manual} & \multicolumn{2}{c}{ \specialcell{{\it Movie$\rightarrow$Actor$\rightarrow$Movie}, {\it Movie$\rightarrow$Director$\rightarrow$Movie}}} \\
    % \midrule
    \hline
    \specialcell{GTN~\cite{YJKKK19}} & \multicolumn{2}{c}{ \specialcell{{\it Movie$\rightarrow$ Director}, {\it Movie$\rightarrow$ Actor} \\ {\it Movie$\rightarrow$Director$\rightarrow$Movie}} } \\
    % \bottomrule
    \hline
\end{tabular}
% }
\vspace{-3mm}
\end{table}

\smallskip\noindent
\textbf{Overall Meta-path Statistics.}
We further present meta-paths generated for most of the target nodes in Table~\ref{table:meta_paths_IMDb}\footnote{We do not present the relation types for all meta-paths in Table~\ref{table:meta_paths_IMDb}.}, which are summarised from Fig.~\ref{fig:action_analysis_IMDb}.
Table~\ref{table:meta_paths_DBLP} and Table~\ref{table:meta_paths_ACM} present the generated meta-paths on DBLP and ACM, respectively.
We can find from Table~\ref{table:meta_paths_IMDb} that the RL agent can generate the same meta-paths as manually-defined ones.
Besides, we find that these meta-paths specified by human experts are not the most useful ones to learn node representations for {\it Movie} nodes.
Two short meta-paths ${\it Movie} \xrightarrow{\it MA} {\it Actor}$ and ${\it Movie} \xrightarrow{\it MD} {\it Director}$ are the most useful ones.
This indicates that participants (director and actors) of the movie largely determine its type as the task is to predict the class of a movie.
The top three meta-paths generated by the most potent competing method, GTN, confirm our observation that two shorter meta-paths are more valuable to learn {\it Movie} nodes' representations.
However, GTN can only select several meta-paths for every node type; our model can identify personalised meta-paths for each node. 
More analyses on Table~\ref{table:meta_paths_DBLP} and Table~\ref{table:meta_paths_ACM} can be found in Appendix~\ref{sec:appendix_more_meta-paths_analysis}. 

\begin{figure}[!ht]
\centering
\includegraphics[width=.5\linewidth]{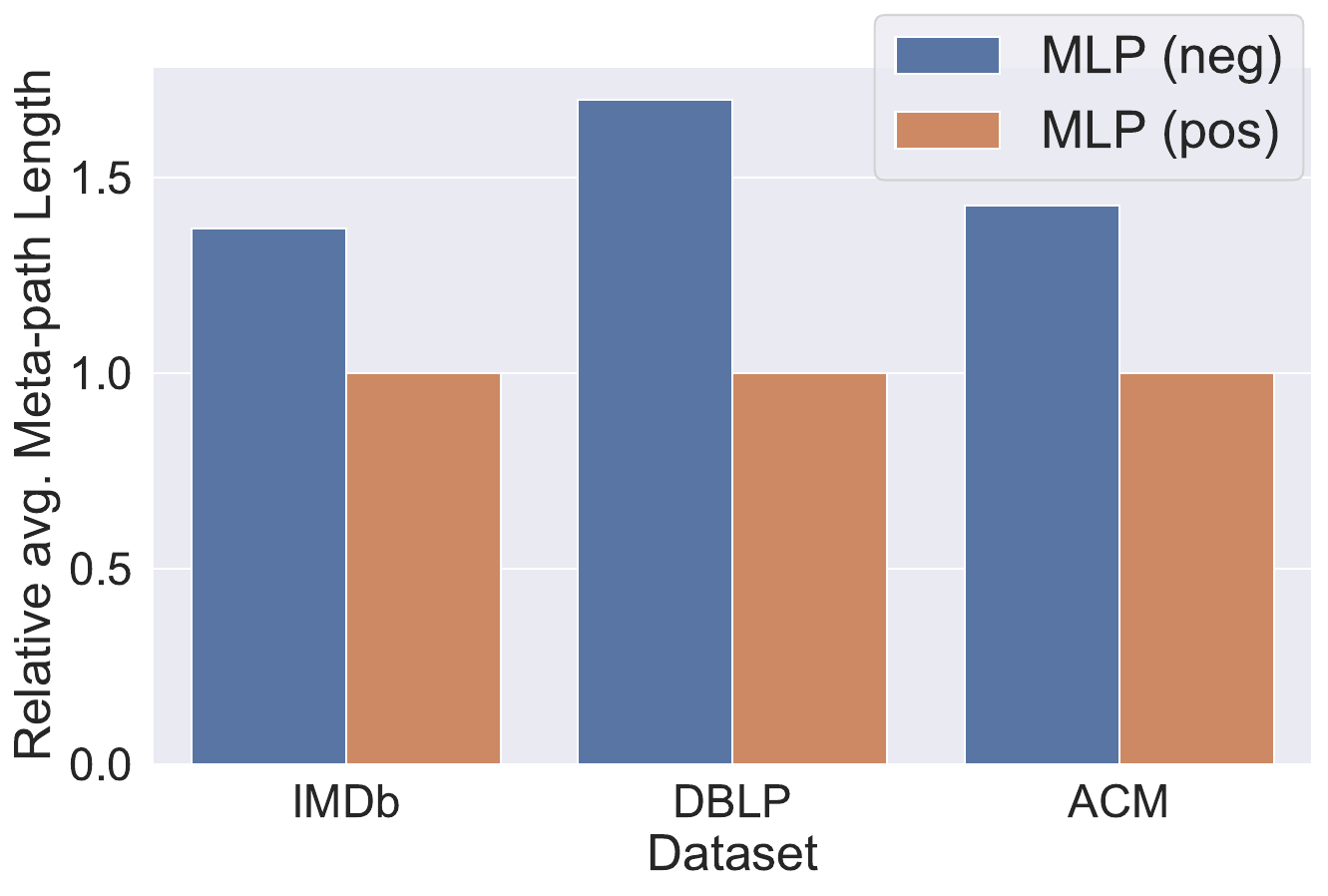}
\caption{
    The average meta-path length generated by PM-HGNN\textit{++} on different datasets. 
    MLP (pos)/ MLP (neg) are groups of nodes that MLP makes correct and wrong classifications, respectively.
}
\label{fig:meta-path_length_analysis}
\vspace{-3mm}
\end{figure}

\smallskip\noindent
\textbf{Meta-path Comparison.}
In order to understand how does PM-HGNN\textit{++} generate personalised meta-paths for different nodes, we hereby present another analysis to investigate the designed meta-path lengths for nodes that can be correctly classified according to raw node attributes.
Specifically, we devide nodes into two groups, \textit{pos} and \textit{neg}, according to whether applying multi-layer perceptron (MLP) with node attributes can produce correct classification under supervised settings. 
That said, correctly-classified nodes and incorrectly-classified ones are grouped into \textit{pos} and \textit{neg}, respectively.
Then, we calculate the average lengths of PM-HGNN\textit{++} generated meta-paths for nodes belonging to two groups. 
The results are reported in Fig.~\ref{fig:meta-path_length_analysis}. 
We can find that nodes that cannot be correctly classified by MLP are associated with longer personalised meta-paths. 
Such results deliver an important insight: PM-HGNN\textit{++} tends to explore deeper relational neighbouring structure to enhance node representation learning, i.e., discriminating nodes with different labels from one another, if node attributes themselves cannot provide sufficient information. 
Moreover, in the DBLP dataset, which has more complicated heterogeneous semantic patterns (4 node types and 6 relation types), node group MLP (\textit{neg}) has relatively longer meta-paths because PM-HGNN\textit{++} can have more options to explore and generate personalised meta-paths. 

\subsection{Model Analysis on PM-HGNN\textit{++}}
\label{subsec:model_analysis_study}

\begin{figure}[!ht]
\centering
\includegraphics[width=.4\linewidth]{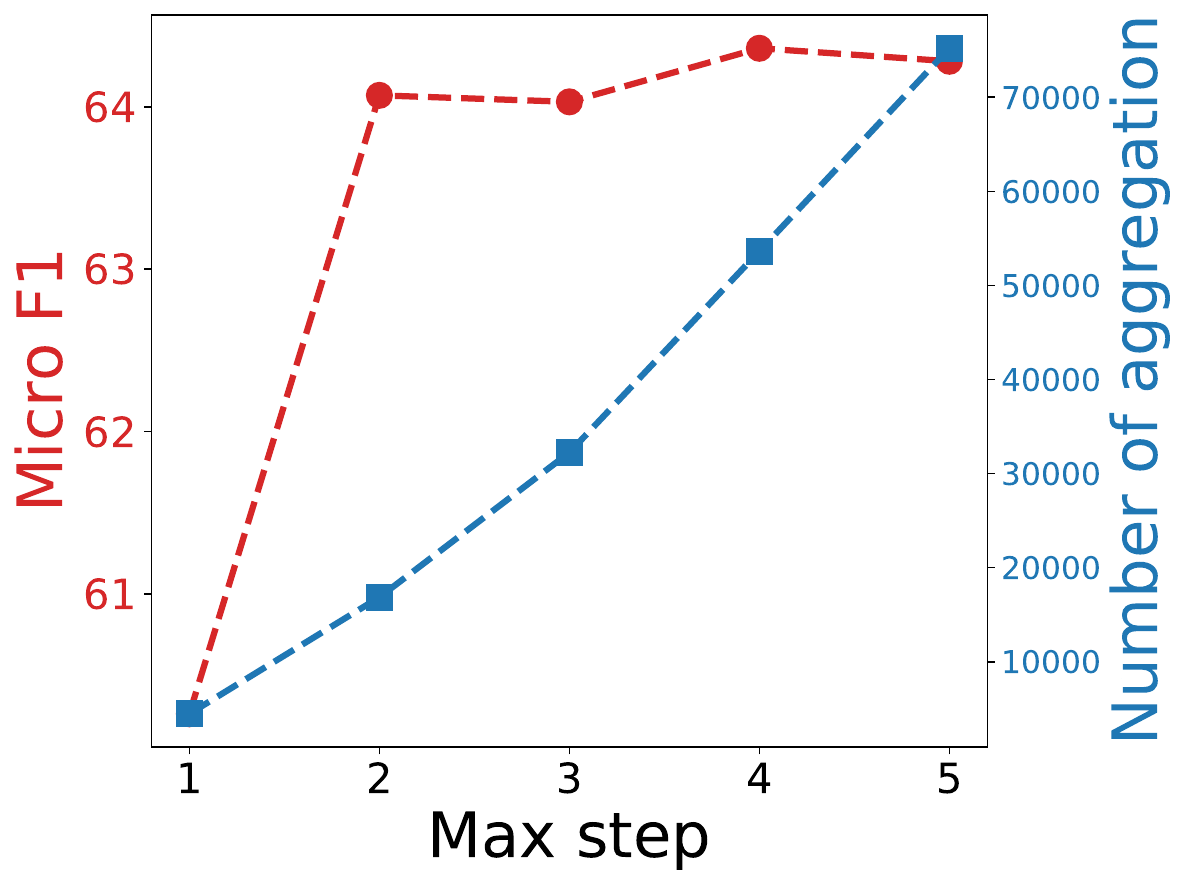}
\caption{Micro-F1 and the number of aggregations of PM-HGNN\textit{++} on IMDb, with different max steps.}
\label{fig:max_timestep_analysis}
\centering
\vspace{-3mm}
\end{figure}

\noindent
\textbf{Maximum Step $T$.}
We study how the influence of maximum step $T$ and the number of aggregation paths affect on the classification performance.
The results on IMDb dataset are presented in Fig.~\ref{fig:max_timestep_analysis}.
Setting $T \geq 2$ leads to the most significant performance improvement in terms of Micro F1 over $T\!=\!1$.
As $T$ increases, the number of aggregation paths also increases.
Considering that more aggregation paths bring higher computational cost, we choose $T\!=\!2$ for the experiments in Sec.~\ref{subsec:experimental_results}, even though $T\!>\!2$ can produce better performance.

\begin{figure}[!ht]
\centering
\includegraphics[width=.75\linewidth]{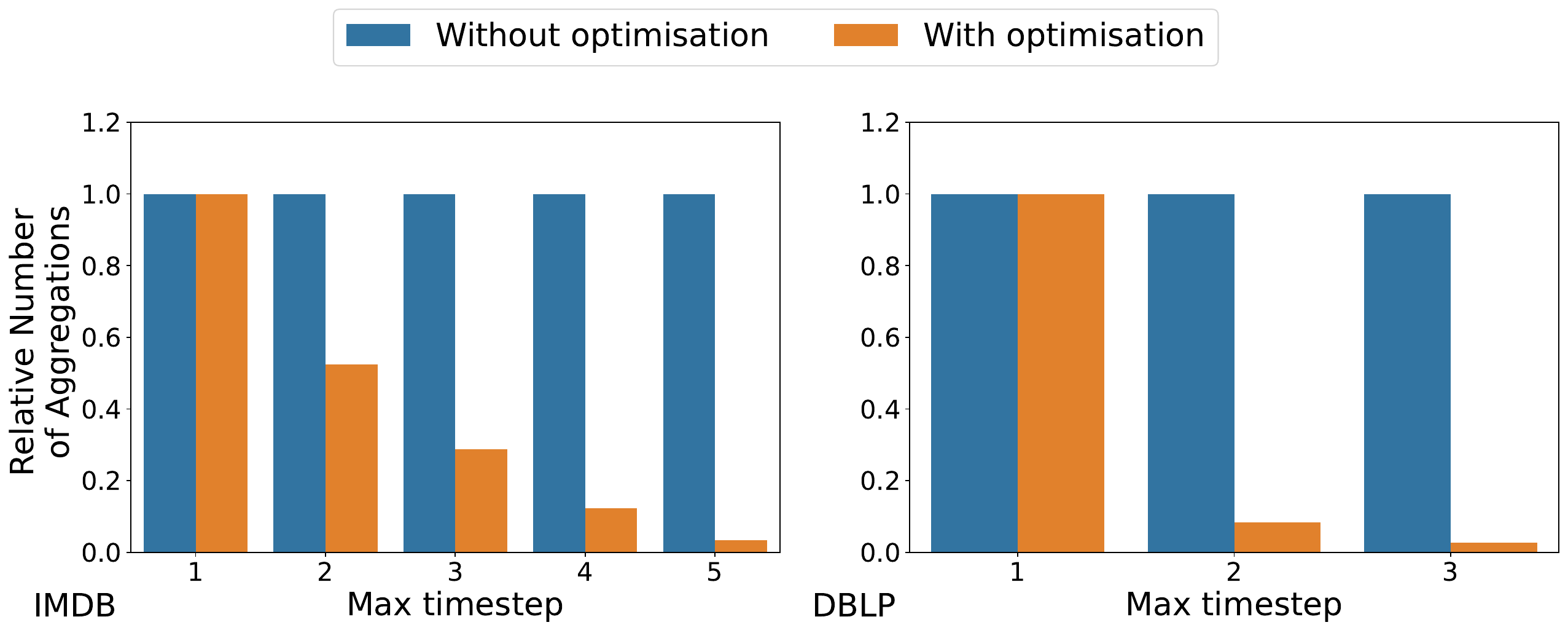}
\caption{
    The relative numbers of aggregations (\# of without the redundancy-free aggregation divided by \# of with the redundancy-free aggregation) under different maximum steps on IMDb (left) and DBLP (right). 
}
\label{fig:redundancy_analysis}
\vspace{-3mm}
\end{figure}

\smallskip\noindent
\textbf{Redundancy-free Aggregation.}
We investigate the influence of the redundancy-free improvement according to the number of aggregations with/without the improvement.
Fig.~\ref{fig:redundancy_analysis} shows the results.
We can find that our redundancy-free aggregation strategy significantly reduces the number of information aggregations.
Besides, as the maximum step $T$ increases, the reduction effect is more obvious.
On IMDb, our method reduces near 50\% aggregations with $T=2$, and the reduced ratio is more than 80\% when $T=4$.

\begin{figure}[!ht]
\centering
\includegraphics[width=.75\linewidth]{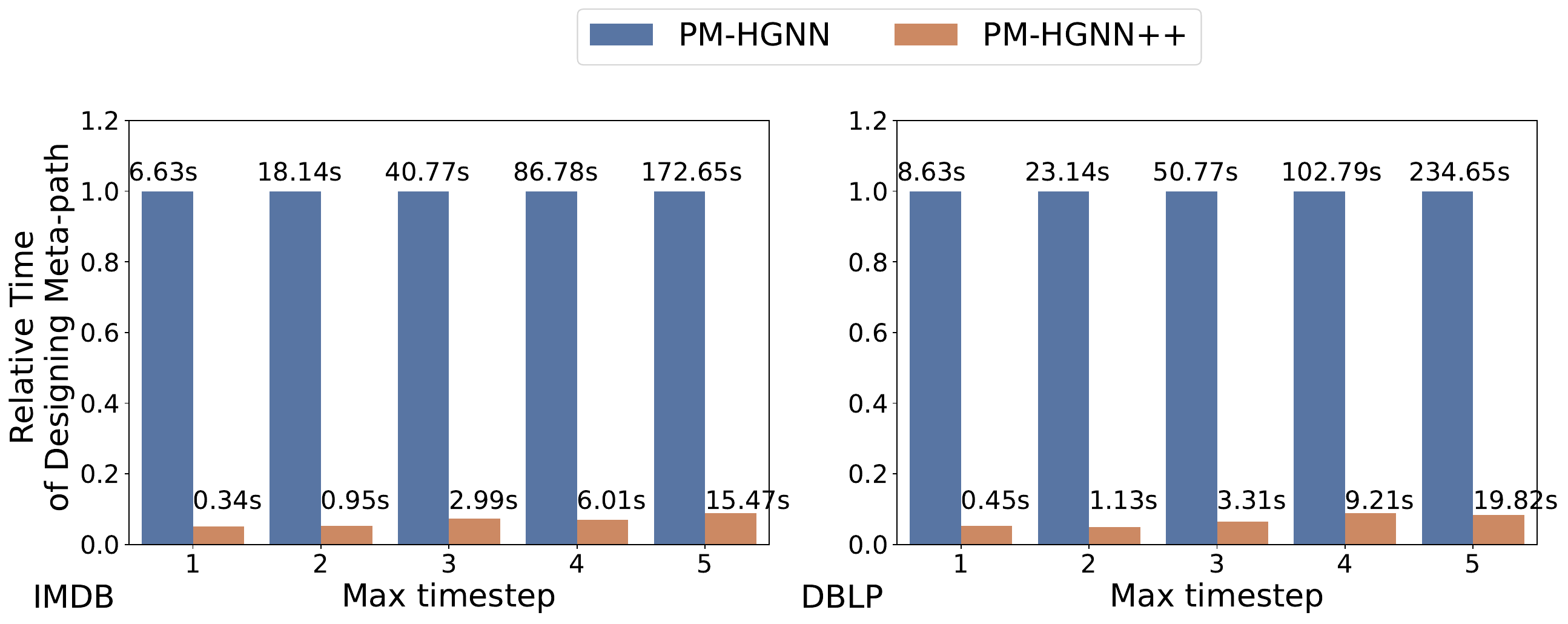}
\caption{
    Run time analysis of meta-path generation for PM-HGNN and PM-HGNN\textit{++} under different max steps based on IMDb (left) and DBLP (right) datasets. The x-axis is the number of maximum timestep in generating meta-paths, and the y-axis is the relative time (i.e., the run time of PM-HGNN\textit{++} divided by the run time of PM-HGNN). Besides, the numbers at tops of bars indicate the run time values in seconds.
}
\label{fig:run_time_analysis}
\vspace{-3mm}
\end{figure}

\smallskip\noindent
\textbf{Run Time Analysis of Meta-path Generation.}
As described in Sec.~\ref{subsec:PM-HGNN_++}, PM-HGNN\textit{++} can accelerate the meta-path generation process through the proposed novel training process.
Here we compare the run time of PM-HGNN and PM-HGNN\textit{++} for personalised meta-paths generation.
We report their actual run time in seconds, and calculate the run time of PM-HGNN\textit{++} relative to PM-HGNN (i.e., relative time). The results on IMDb and DBLP datasets are shown in Fig.~\ref{fig:run_time_analysis}.
We can find that PM-HGNN\textit{++} is less than $5\%$ of PM-HGNN's run time on both datasets. 
Such results exhibit the promising time efficiency of PM-HGNN\textit{++}.

% section experiments (end)
%------------------------------------------------------------------------------

%------------------------------------------------------------------------------
\section{Conclusions and future work} % (fold)
\label{sec:conclusion_and_future_work}
We have studied in this paper the HGRL problem and identified the limitation of existing HGRL methods, i.e., mainly due to their dependency on hand-crafted meta-paths.
In order to fully unleash the power of HGRL, we presented a novel framework PM-HGNN and proposed one extension model PM-HGNN\textit{++}.
Compared with existing HGRL models, the most significant advantages of our framework lie in avoiding manual efforts in defining meta-paths of HGRL and generating personalised meta-paths for each individual node.
The experimental results demonstrated that our framework generally outperforms the competing approaches and discovers useful meta-paths that have been ignored by human expertise.
In the future, we plan to extend our framework to other tasks on HINs, such as online recommendation and knowledge graph completion and understanding PM-HGNN's generated meta-paths is another promising direction. 

% section conclusion_and_future_work (end)
%------------------------------------------------------------------------------

\bmhead{Acknowledgments}
This work is supported by the Luxembourg National Research Fund through grant PRIDE15/10621687/SPsquared, and supported by Ministry of Science and Technology (MOST) of Taiwan under grants 110-2221-E-006-136-MY3, 110-2221-E-006-001, and 110-2634-F-002-051.

% \bibliography{sn-bibliography}% common bib file
\bibliography{full_format_references}


\clearpage
\appendix
%------------------------------------------------------------------------------
%------------------------------------------------------------------------------
\section{Datasets}
\label{sec:appendix_datasets}
The statistics of datasets are summarised in Table~\ref{table:dataset}. 
Detailed descriptions of two datasets are presented as follows:

\smallskip\noindent
\textbf{IMDb}\footnote{\url{https://www.imdb.com/}} is an online dataset about movies and television programs, including information such as cast, production crew and plot summaries.
We extract a subset of IMDb that contains $4,278$ movies, $2,081$ directors and $5,257$ actors.
The movies are labelled as one of the three classes, i.e., \textit{Action}, \textit{Comedy} and \textit{Drama}, according to their genre information.
The attribute of each movie corresponds to elements of a bag of words (i.e., their plot keywords, $1,232$ in total).

\smallskip\noindent
\textbf{DBLP}\footnote{\url{https://dblp.uni-trier.de/}} is an online computer science bibliography.
We extract a subset of DBLP that contains $4,057$ movies, $1,4328$ papers, $7,723$ terms and $20$ venues.
The authors are labelled as one of the following four research areas: \textit{Database}, \textit{Data mining}, \textit{Machine learning} and \textit{Information retrial}.
Each author can be described by a bag of words (i.e., their paper keywords, $334$ in total).

\smallskip\noindent
\textbf{ACM}\footnote{\url{https://dl.acm.org/}} is an online academic publication dataset. 
We extract papers published in KDD, SIGMOD, SIGCOMM, MobiCOMM and VLDB and divided the papers into three classes (\textit{Database}, \textit{Wireless Communication}, and \textit{Data Mining}). 
Then, we construct a HIN that comprises $3,025$ papers, $5,835$ authors and $56$ subjects.
Paper features correspond to elements of a bag of words represented by keywords.
We label the papers according to the conference they published. 

\begin{table}[ht!]
\caption{Statistics of the datasets.}
% \vspace{-3mm}
\label{table:dataset}
% \small
% \scriptsize
% \resizebox{.7\linewidth}{!}{
\begin{tabular}{l | c | c | c }
    \hline
    Dataset & Node & Relation & \# Attributes \\
    \hline
    IMDb & \specialcell{\# Movie (M): 4,278 \\ \# Director (D): 2,081 \\ \# Actor (A): 5,257} & \specialcell{\# M-D, \# D-M: 4,278 \\ \# M-A, \# A-M: 12,828} & 1,232 \\
    \hline
    DBLP & \specialcell{\# Author (A): 4,057 \\ \# Paper (P): 14,328 \\ \# Term (T): 7,723 \\ \# Venue (V): 20} & \specialcell{\# A-P \# P-A: 19,645 \\ \# P-T \# T-P: 85,810 \\ \# P-V \# V-P: 14,328} & 334 \\
    \hline
    ACM & \specialcell{\# Paper (P): 3,025 \\ \# Author (A): 5,835 \\ \# Subject (S): 56} & \specialcell{\# A-P \# P-A: 9,744 \\ \# P-S \# S-P: 3,025 } & 1,830 \\
    \hline
\end{tabular}
% }
\vspace{-3mm}
\end{table}

\begin{figure}[!ht]
\centering
\includegraphics[width=.7\linewidth]{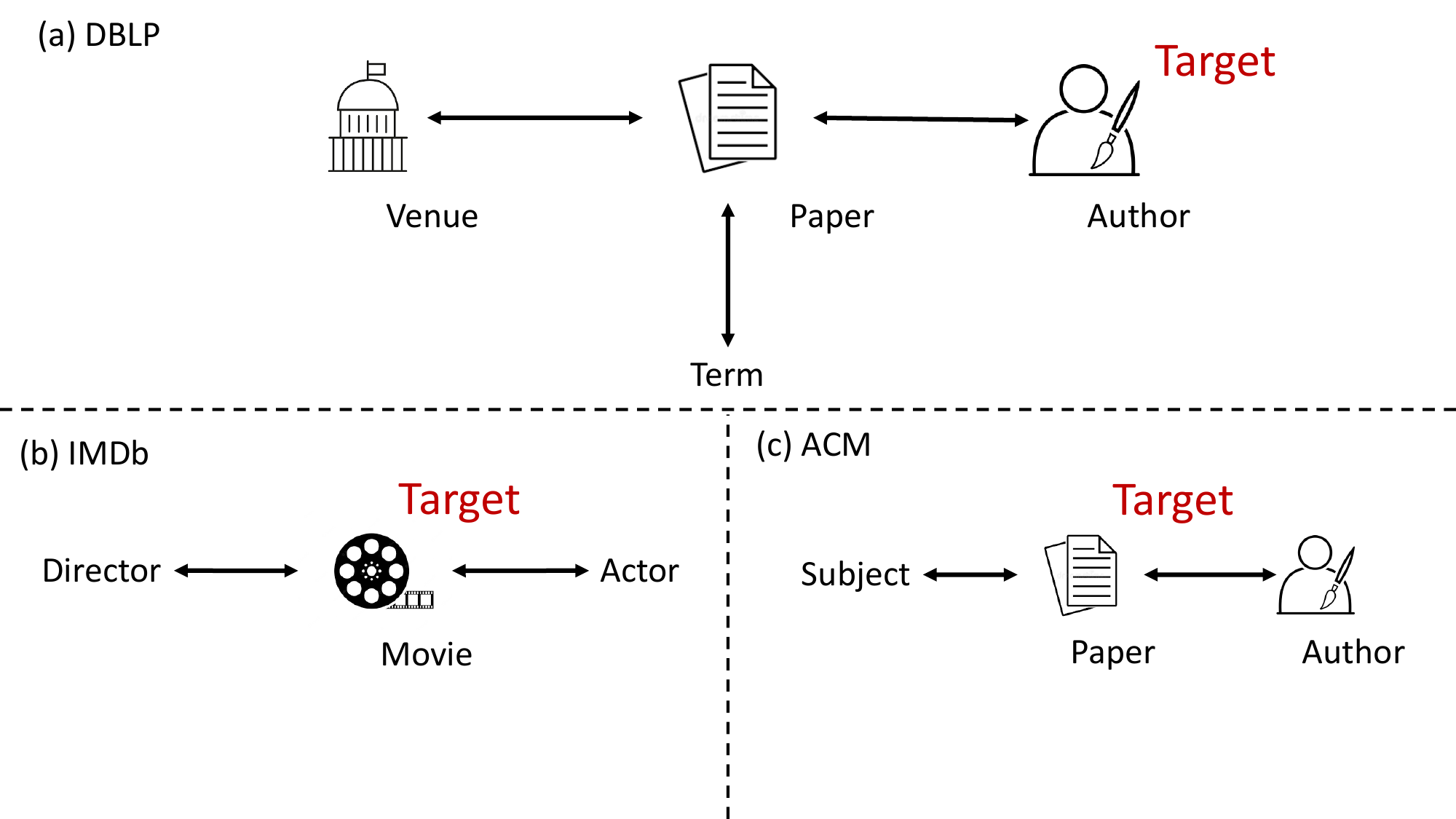}
\caption{
    Meta-relation schema of three datasets. 
}
\label{fig:meta-relation_schema}
% \vspace{-3mm}
\end{figure}

%------------------------------------------------------------------------------

%------------------------------------------------------------------------------
\section{More Meta-paths Analysis}
\label{sec:appendix_more_meta-paths_analysis}
\begin{figure}[!ht]
\centering
\includegraphics[width=.7\linewidth]{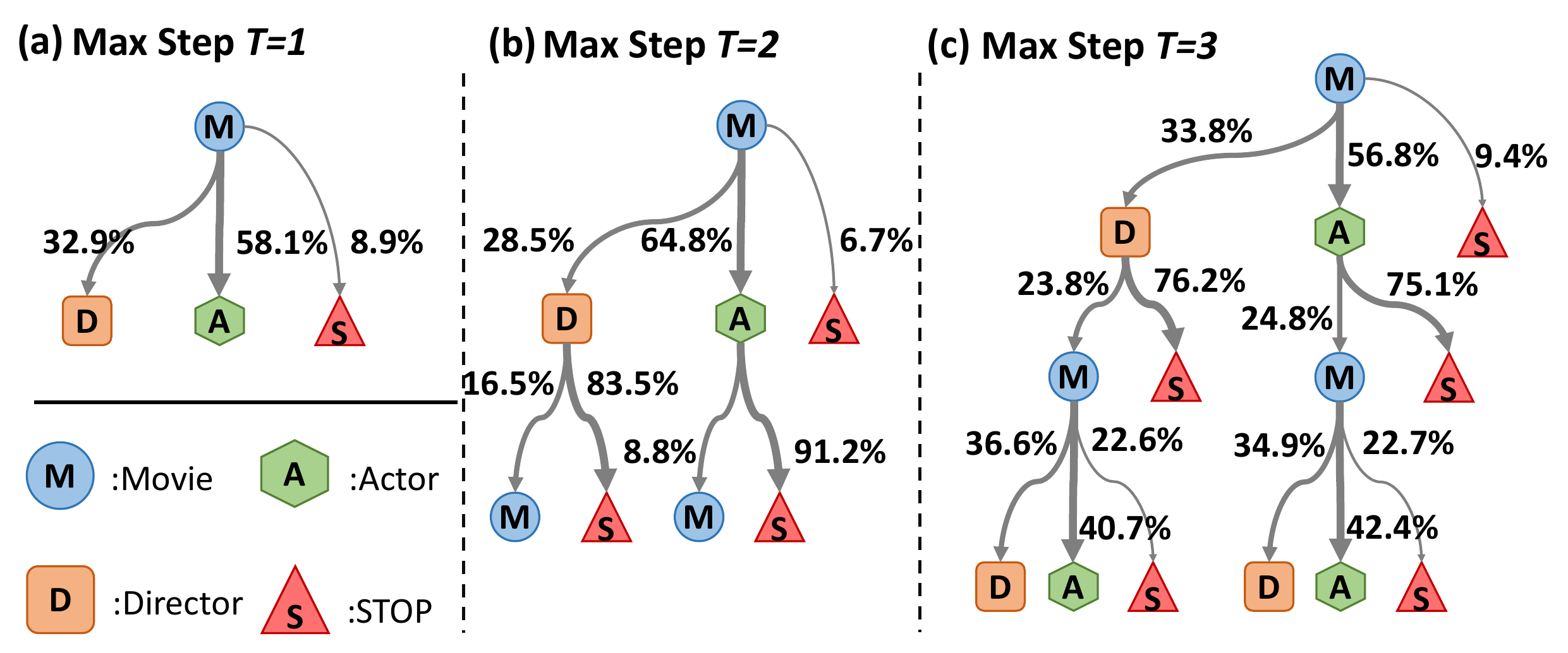}
\caption{
    Actions that the RL agent takes on IMDb: (a), (b) and (c) correspond to PM-HGNN\textit{++} with max step $T=1, 2, 3$, respectively. 
    The red triangles with ``S'' indicate the ``STOP'' action.
    The thickness of links represents the ratio of the corresponding action.
}
\label{fig:action_analysis_IMDb}
% \vspace{-3mm}
\end{figure}

\begin{table}[!ht]
\caption{
All possible meta-paths on different datasets.
IMDb: Movie (M), Director (D), Actor (A);
DBLP: Author (A), Paper (P), Term (T), Venue (V);
ACM: Paper (P), Author (A), Subject (S).
}
% \vspace{-3mm}
\label{table:all_possible_meta_paths}
% \scriptsize
\centering
% \resizebox{.65\linewidth}{!}{
\begin{tabular}{c|c|c|c}
    \hline
    $t$     & IMDb & DBLP & ACM \\ 
    \hline
    $t$ = 1 
            & \specialcell{ {\it M$\rightarrow$A} \\ {\it M$\rightarrow$D} } 
            & \specialcell{ {\it A$\rightarrow$P} } 
            & \specialcell{ {\it P$\rightarrow$S} \\ {\it P$\rightarrow$A} } \\ 
    \hline
    $t$ = 2 
            & \specialcell{ {\it M$\rightarrow$A$\rightarrow$M} \\ {\it M$\rightarrow$D$\rightarrow$M} } 
            & \specialcell{ {\it A$\rightarrow$P$\rightarrow$A} \\ {\it A$\rightarrow$P$\rightarrow$V} \\ {\it A$\rightarrow$P$\rightarrow$T} } 
            & \specialcell{ {\it P$\rightarrow$S$\rightarrow$P} \\ {\it P$\rightarrow$A$\rightarrow$P} } \\ 
    \hline
    $t$ = 3 
            & \specialcell{ {\it M$\rightarrow$A$\rightarrow$M$\rightarrow$A} \\ {\it M$\rightarrow$A$\rightarrow$M$\rightarrow$D} \\ {\it M$\rightarrow$D$\rightarrow$M$\rightarrow$A} \\ {\it M$\rightarrow$D$\rightarrow$M$\rightarrow$D} } 
            & \specialcell{ {\it A$\rightarrow$P$\rightarrow$A$\rightarrow$P} \\ {\it A$\rightarrow$P$\rightarrow$V$\rightarrow$P} \\ {\it A$\rightarrow$P$\rightarrow$T$\rightarrow$P} } 
            & \specialcell{ {\it P$\rightarrow$S$\rightarrow$P$\rightarrow$S} \\ {\it P$\rightarrow$S$\rightarrow$P$\rightarrow$A} \\ {\it P$\rightarrow$A$\rightarrow$P$\rightarrow$S} \\ {\it P$\rightarrow$A$\rightarrow$P$\rightarrow$A} } \\ 
    \hline
    $t$ = 4 
            & \specialcell{ {\it M$\rightarrow$A$\rightarrow$M$\rightarrow$A$\rightarrow$M} \\ {\it M$\rightarrow$A$\rightarrow$M$\rightarrow$D$\rightarrow$M} \\ {\it M$\rightarrow$D$\rightarrow$M$\rightarrow$A$\rightarrow$M} \\ {\it M$\rightarrow$D$\rightarrow$M$\rightarrow$D$\rightarrow$M} } 
            & \specialcell{ {\it A$\rightarrow$P$\rightarrow$A$\rightarrow$P$\rightarrow$A} \\ {\it A$\rightarrow$P$\rightarrow$A$\rightarrow$P$\rightarrow$V} \\ {\it A$\rightarrow$P$\rightarrow$A$\rightarrow$P$\rightarrow$T} \\ {\it A$\rightarrow$P$\rightarrow$V$\rightarrow$P$\rightarrow$A} \\ {\it A$\rightarrow$P$\rightarrow$V$\rightarrow$P$\rightarrow$V} \\ {\it A$\rightarrow$P$\rightarrow$V$\rightarrow$P$\rightarrow$T} \\ {\it A$\rightarrow$P$\rightarrow$T$\rightarrow$P$\rightarrow$A} \\ {\it A$\rightarrow$P$\rightarrow$T$\rightarrow$P$\rightarrow$V} \\ {\it A$\rightarrow$P$\rightarrow$T$\rightarrow$P$\rightarrow$T} } 
            & \specialcell{ {\it P$\rightarrow$S$\rightarrow$P$\rightarrow$S$\rightarrow$P} \\ {\it P$\rightarrow$S$\rightarrow$P$\rightarrow$A$\rightarrow$P} \\ {\it P$\rightarrow$A$\rightarrow$P$\rightarrow$S$\rightarrow$P} \\ {\it P$\rightarrow$A$\rightarrow$P$\rightarrow$A$\rightarrow$P} } \\ 
    \hline
\end{tabular}
% }
% \vspace{-3mm}
\end{table}

We have introduced three datasets that we used in the experimental sections in Section~\ref{subsec:experimental_settings} and Section~\ref{sec:appendix_datasets}. 
Here we further present the set of meta-paths that are possibly generated in each step in Table~\ref{table:all_possible_meta_paths}\footnote{We only discuss the meta-paths that started with target nodes of each dataset.}. 
It should be noted that as we discussed in Section~\ref{sec:methodology}, there is always an available \textit{STOP} action at each step.
That said, PM-HGNN variants allow to design meta-paths with flexible lengths at each step. 

\begin{figure}[!ht]
\centering
\includegraphics[width=.7\linewidth]{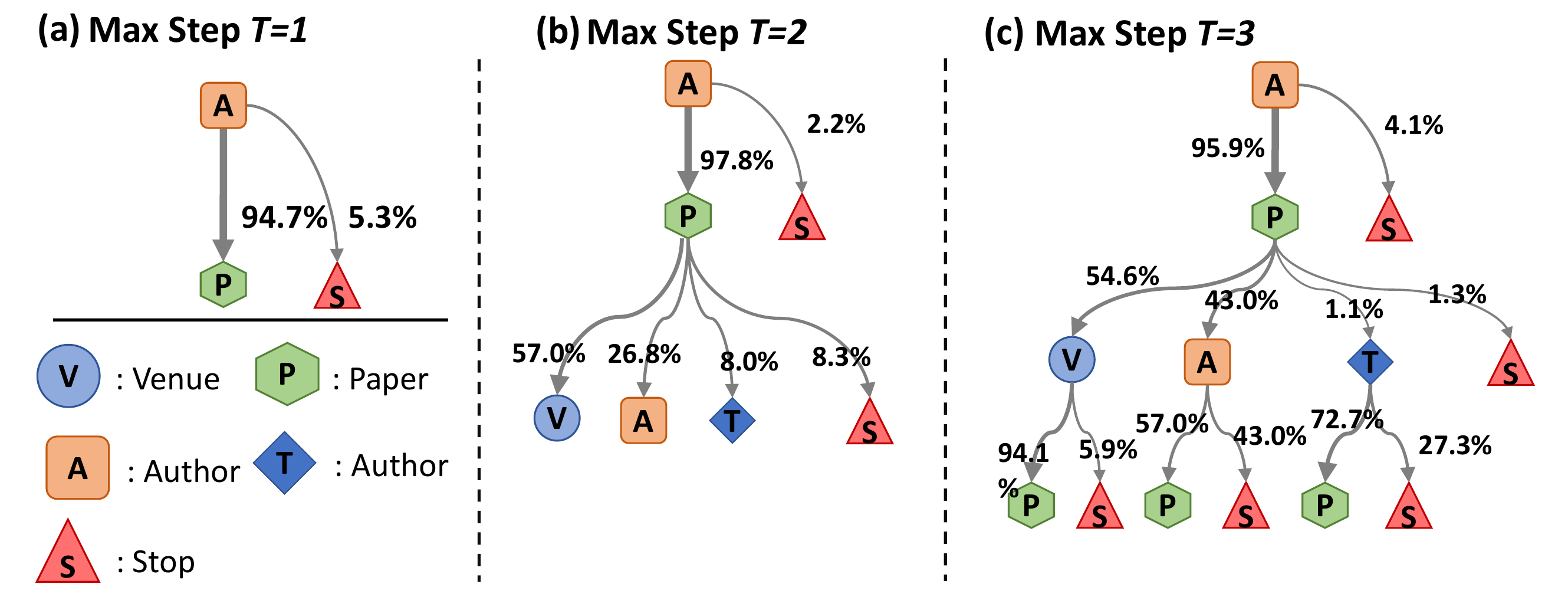}
\caption{
    Actions that the RL agent takes on DBLP: (a), (b) and (c) correspond to PM-HGNN\textit{++} with max step $T=1, 2, 3$, respectively. 
    The red triangles with ``S'' indicate the ``STOP'' action.
    The thickness of links represents the ratio of the corresponding action.
}
\label{fig:action_analysis_DBLP}
% \vspace{-3mm}
\end{figure}

\begin{table}[!ht]
\caption{Meta-paths generated by PM-HGNN\textit{++} on DBLP.}
% \vspace{-3mm}
\label{table:meta_paths_DBLP}
% \scriptsize
\centering
% \resizebox{.65\linewidth}{!}{
\begin{tabular}{c|c|c}
    % \toprule
    \hline
    $T$                 & \specialcell{Meta-paths \\ Designed by PM-HGNN\textit{++}} & Percentage (\%) \\ 
    % \midrule
    \hline
    $T$ = 1             & \specialcell{ {\it Author$\rightarrow$Paper} } & \specialcell{94.7} \\ 
    % \midrule
    \hline
    $T$ = 2             & \specialcell{ {\it Author$\rightarrow$Paper$\rightarrow$Venue} \\ 
    {\it Author$\rightarrow$Paper$\rightarrow$Author} \\ 
    {\it Author$\rightarrow$Paper} \\ 
    {\it Author$\rightarrow$Paper$\rightarrow$Term} } & \specialcell{55.7 \\ 26.2 \\ 8.1 \\ 7.8} \\ 
    % \midrule
    \hline
    $T$ = 3             & \specialcell{ {\it Author$\rightarrow$Paper$\rightarrow$Venue$\rightarrow$Paper} \\ 
    {\it Author$\rightarrow$Paper$\rightarrow$Author$\rightarrow$Paper} \\ 
    {\it Author$\rightarrow$Paper$\rightarrow$Author} \\ 
    {\it Author$\rightarrow$Paper$\rightarrow$Venue} \\ 
    {\it Author$\rightarrow$Paper} } & \specialcell{49.3 \\ 23.5 \\ 17.7 \\ 3.1 \\ 1.2} \\ 
    \hline
    $T$ = 4             & \specialcell{ {\it Author$\rightarrow$Paper$\rightarrow$Venue$\rightarrow$Paper$\rightarrow$Author} \\
    {\it Author$\rightarrow$Paper$\rightarrow$Author} \\ 
    {\it Author$\rightarrow$Paper$\rightarrow$Author$\rightarrow$Paper$\rightarrow$Author} \\ 
    {\it Author$\rightarrow$Paper$\rightarrow$Venue$\rightarrow$Paper} \\ 
    {\it Author$\rightarrow$Paper$\rightarrow$Venue} } & \specialcell{32.7 \\ 25.8 \\ 21.9 \\ 14.9 \\ 3.2} \\ 
    \hline
    \specialcell{Manual} & \multicolumn{2}{c}{ \specialcell{ {\it Author$\rightarrow$Paper$\rightarrow$Venue$\rightarrow$Paper$\rightarrow$Author}, \\ 
    {\it Author$\rightarrow$Paper$\rightarrow$Author}, \\
    {\it Author$\rightarrow$Paper$\rightarrow$Term$\rightarrow$Paper$\rightarrow$Author} } } \\
    \hline
    \specialcell{GTN~\cite{YJKKK19}} & \multicolumn{2}{c}{ \specialcell{{\it Author$\rightarrow$Paper$\rightarrow$Venue$\rightarrow$Paper$\rightarrow$Author}, \\ {\it Author$\rightarrow$Paper$\rightarrow$Author$\rightarrow$Paper$\rightarrow$Author}, \\
    {\it Author$\rightarrow$Paper$\rightarrow$Author} } } \\
    % \bottomrule
    \hline
\end{tabular}
% }
% \vspace{-3mm}
\end{table}

\begin{figure}[!ht]
\centering
\includegraphics[width=.7\linewidth]{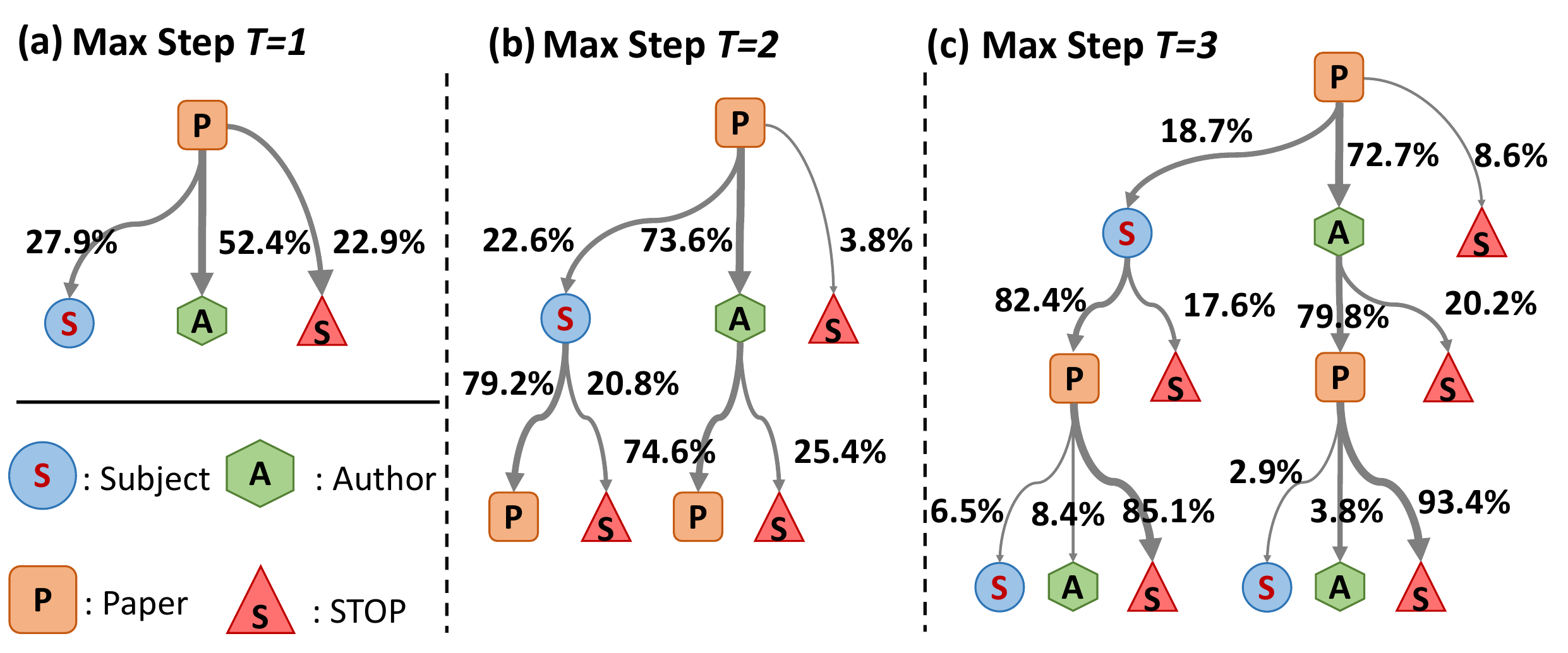}
\caption{
    Actions that the RL agent takes on ACM: (a), (b) and (c) correspond to PM-HGNN\textit{++} with max step $T=1, 2, 3$, respectively. 
    The red triangles with a black ``S'' indicate the ``STOP'' action.
    The thickness of links represents the ratio of the corresponding action.
}
\label{fig:action_analysis_ACM}
% \vspace{-3mm}
\end{figure}

\begin{table}[!ht]
\caption{Meta-paths generated by PM-HGNN\textit{++} on ACM.}
% \vspace{-3mm}
\label{table:meta_paths_ACM}
% \scriptsize
\centering
% \resizebox{.65\linewidth}{!}{
\begin{tabular}{c|c|c}
    % \toprule
    \hline
    $T$                 & \specialcell{Meta-paths \\ Designed by PM-HGNN\textit{++}} & Percentage (\%) \\ 
    % \midrule
    \hline
    $T$ = 1             & \specialcell{ {\it Paper$\rightarrow$Author} \\ {\it Paper$\rightarrow$Subject} } & \specialcell{52.4 \\ 27.9} \\ 
    % \midrule
    \hline
    $T$ = 2             & \specialcell{ {\it Paper$\rightarrow$Author$\rightarrow$ Paper} \\ {\it Paper$\rightarrow$Author} \\ {\it Paper $\rightarrow$Subject$\rightarrow$Paper} \\ {\it Paper$\rightarrow$Subject} } & \specialcell{ 54.9 \\ 18.7 \\ 17.9 \\ 4.7 } \\ 
    % \midrule
    \hline
    $T$ = 3             & \specialcell{ {\it Paper$\rightarrow$Author$\rightarrow$Paper} \\ {\it Paper$\rightarrow$Author} \\ {\it Paper$\rightarrow$Subject$\rightarrow$Paper} \\ {\it Paper$\rightarrow$Subject} \\ {\it Paper $\rightarrow$Author$\rightarrow$Paper$\rightarrow$Author} } & \specialcell{54.2 \\ 14.7 \\ 13.1 \\ 3.3 \\ 2.1} \\ 
    % \toprule
    \hline
    \specialcell{Manual} & \multicolumn{2}{c}{ \specialcell{ {\it Paper$\rightarrow$Author$\rightarrow$Paper}, {\it Paper$\rightarrow$Subject$\rightarrow$Paper}} } \\
    % \midrule
    \hline
    \specialcell{GTN~\cite{YJKKK19}} & \multicolumn{2}{c}{ \specialcell{ {\it Paper$\rightarrow$Author$\rightarrow$Paper}, {\it Paper$\rightarrow$Subject$\rightarrow$Paper}} } \\
    % \bottomrule
    \hline
\end{tabular}
% }
% \vspace{-3mm}
\end{table}

We present more meta-path analysis here. 
Figure~\ref{fig:action_analysis_DBLP}\footnote{We present the visualisation results of $T<4$, since the figure will be too dense to see the content clearly with $T=4$} and Figure~\ref{fig:action_analysis_ACM} visualise how the reinforcement learning agent in PM-HGNN\textit{++} generates personalised meta-paths for each target node on DBLP and ACM datasets with different max steps under the semi-supervised setting, respectively 
The percentages marked in the figure represent the fraction of nodes choosing the corresponding relation to extend the meta-path at that step.
Table~\ref{table:meta_paths_DBLP} and Table~\ref{table:meta_paths_ACM} summarise the predefined meta-paths, the found important meta-paths by GTN~\cite{YJKKK19} and the top-frequent personalised meta-paths designed by PM-HGNN\textit{++} for the target nodes of DBLP and ACM datasets, respectively. 
The meta-path generation process is shown in Figure~\ref{fig:action_analysis_DBLP} and Figure~\ref{fig:action_analysis_ACM}, the RL agent makes decisions to extend the meta-paths for different nodes according to the defined state $S$. 
Note that, we only present the top-$5$ frequent meta-paths if there are more possibilities. 

%------------------------------------------------------------------------------

%------------------------------------------------------------------------------
\section{Competing Methods}
\label{sec:appendix_competing_methods}
We adopt $5$ homogeneous graph representation learning models and $10$ heterogeneous graph representation learning models and $1$ state-of-the-art relational learning model as competing methods.

\begin{enumerate}
% \smallskip\noindent
\item
\textbf{LINE}~\cite{TQWZYM15} is a traditional homogeneous model that exploits the first-order and second-order proximity between nodes.
We apply it to the datasets by ignoring heterogeneous information network heterogeneity and node attributes.

% \smallskip\noindent
\item
\textbf{DeepWalk}~\cite{PAS14} is a random walk-based graph representation learning method for homogeneous graphs, we apply it to the datasets by ignoring heterogeneous information network heterogeneity and node attributes.

% \smallskip\noindent
\item
\textbf{ESim}~\cite{SQLKHP16} is a heterogeneous graph representation learning method that learns different semantics from multiple meta-paths.
It requires a pre-defined weight for each meta-path, we assign all meta-paths with the same weight.

% \smallskip\noindent
\item
\textbf{metapath2vec}~\cite{DCS17} is a heterogeneous graph representation learning approach that performs random walks on heterogeneous information networks with the guidance of pre-defined meta-paths and utilises skip-gram to generate node representations.

% \smallskip\noindent
\item
\textbf{JUST}~\cite{HYC18} is a heterogeneous graph representation learning method that does not rely on manually defined meta-paths.

% \smallskip\noindent
\item
\textbf{HERec}~\cite{SHZY19} is a heterogeneous graph representation learning method that designs a type constraint strategy to filter the node sequence and utilises skip-gram to generate node representations.

% \smallskip\noindent
\item
\textbf{NSHE}~\cite{ZWSLY20}
is a heterogeneous graph representation learning method that tries to maintain network schema during network representation learning.

% \smallskip\noindent
\item
\textbf{MLP}~\cite{H99} is a class of feed-forward neural networks that learns information from node attributes without structural information.

% \smallskip\noindent
\item
\textbf{GCN}~\cite{KW17} is a homogeneous graph neural network that extends the convolution operation to graphs.
Here we test GCN on meta-path-based homogeneous graphs and report the results from the best meta-path.

% \smallskip\noindent
\item
\textbf{GAT}~\cite{VCCRLB18} is a homogeneous graph neural network that performs convolution on graphs with an attention mechanism.
Similar to the implementation of GCN, here we test GAT on meta-path-based homogeneous graphs and report the results from the best meta-path.

% \smallskip\noindent
\item
\textbf{RGCN}~\cite{SKBBTW18} is a heterogeneous graph neural network model which keeps a different weight for each relation to perform convolution on graphs.
The graph neural network encoder is the same as GCN. 

% \smallskip\noindent
\item
\textbf{HAN}~\cite{WJSWYCY19} is a heterogeneous graph neural network model which learns meta-path guided node representations from different meta-path-based homogeneous graphs and integrates them with using attention.

% \smallskip\noindent
\item
\textbf{GTN}~\cite{YJKKK19} is a heterogeneous graph neural network model which transforms a heterogeneous graph into multiple new graphs defined by given meta-paths and selects from them by transition probability. 

\item\textbf{PropStar}~\cite{LSR20} is state-of-the-art relational learning which uses embedding vectors to represent the features of the data set. 
Individual relational features obtained as the result of pro-positionalisation by Wordification, are used by supervised embeddings learners to obtain representations, co-located with instance labels. 

% \smallskip\noindent
\item
\textbf{MAGNN}~\cite{FZMK20} is a heterogeneous graph neural network model which defines meta-path instance encoders to extract the structure and semantics in meta-paths to improve generated representations. 

% \smallskip\noindent
\item
\textbf{HGT}~\cite{HDWS20} is a heterogeneous graph neural network model which uses each meta-relation to parameterise transformer-like self-attention architecture to capture common and specific patterns of relationships. 
\end{enumerate}

For LINE and DeepWalk, we utilise the integrated implementations from GraphEmbedding\footnote{\url{https://github.com/shenweichen/GraphEmbedding}}. 
For metapath2vec, RGCN, HAN and HGT, we utilise the integrated implementations from Deep Graph Library\footnote{\url{https://www.dgl.ai/}}. 
For GCN and GAT, we use the implementation from PytorGeometric\footnote{\url{https://pytorch-geometric.readthedocs.io/en/latest/}}.
For other methods, we adopt the official implementation provided in their published papers: ESim\footnote{\url{https://github.com/shangjingbo1226/ESim}}, JUST\footnote{\url{https://github.com/eXascaleInfolab/JUST}}, HERec\footnote{\url{https://github.com/librahu/HERec}}, NSHE\footnote{\url{https://github.com/AndyJZhao/NSHE}}, GTN\footnote{\url{https://github.com/seongjunyun/Graph_Transformer_Networks}}, PropStart\footnote{\url{https://github.com/SkBlaz/PropStar}} and MAGNN\footnote{\url{https://github.com/cynricfu/MAGNN}}. 

%------------------------------------------------------------------------------

%------------------------------------------------------------------------------
\section{Model Comparison}
\label{sec:appendix_model_comparison}
\begin{table}[ht!]
\caption{
Model comparison from various aspects: Heterogeneous Information Network (HIN), Node-wise Task (NT), End-to-end Training (E2E), Without Manual-defined Meta-paths (WMM), Adaptive Meta-path Generation (AMG), Meta-paths Attached (MPA) to each node-pair ($\ast$), each node type ($\circ$) or each node ($\bullet$).}
% \vspace{-3mm}
\label{table:comparison_diff_models}
\centering
% \small
% \scriptsize
% \footnotesize
% \resizebox{.7\linewidth}{!}{
\begin{tabular}{l | c | c | c | c | c | c}
% \toprule
\hline
& HIN & NT & E2E & WMM & AMG & MPA \\
% \midrule
\hline
\hline
LINE~\cite{TQWZYM15}        &              & $\checkmark$ &              & $\checkmark$ & & \\
% \midrule
\hline
DeepWalk~\cite{PAS14}       &              & $\checkmark$ &              & $\checkmark$ & & \\
% \midrule
\hline
FSPG~\cite{MCMSZ15}         & $\checkmark$ &              &              & $\checkmark$ & $\checkmark$ & $\ast$ \\
% \midrule
\hline
Esim~\cite{SQLKHP16}        & $\checkmark$ & $\checkmark$ &              & $\checkmark$ & & $\circ$ \\
% \midrule
\hline
metapath2vec~\cite{DCS17}   & $\checkmark$ & $\checkmark$ &              & $\checkmark$ & & $\circ$ \\
% \midrule
\hline
JUST~\cite{HYC18}           & $\checkmark$ & $\checkmark$ &              & $\checkmark$ & $\checkmark$ & \\
% \midrule
\hline
HERec~\cite{SHZY19}         & $\checkmark$ & $\checkmark$ &              & $\checkmark$ & & $\circ$ \\
% \midrule
\hline
NSHE~\cite{ZWSLY20}         & $\checkmark$ & $\checkmark$ &              & $\checkmark$ & & $\circ$ \\
% \midrule
\hline
GCN~\cite{KW17}             &              & $\checkmark$ & $\checkmark$ & $\checkmark$ & & \\
% \midrule
\hline
GAT~\cite{VCCRLB18}         &              & $\checkmark$ & $\checkmark$ & $\checkmark$ & & \\
% \midrule
\hline
RGCN~\cite{SKBBTW18}        & $\checkmark$ & $\checkmark$ & $\checkmark$ &              & & $\circ$ \\
% \midrule
\hline
AutoPath~\cite{YLHZPH18}    & $\checkmark$ &              & $\checkmark$ & $\checkmark$ & $\checkmark$ & $\ast$ \\
% \midrule
\hline
HAN~\cite{WJSWYCY19}        & $\checkmark$ & $\checkmark$ & $\checkmark$ &              & & $\circ$ \\
% \midrule
\hline
GTN~\cite{YJKKK19}          & $\checkmark$ & $\checkmark$ & $\checkmark$ &              & $\checkmark$ & $\circ$ \\
% \midrule
\hline
MAGNN~\cite{FZMK20}         & $\checkmark$ & $\checkmark$ & $\checkmark$ &              & & $\circ$ \\
% \midrule
\hline
HGT~\cite{HDWS20}           & $\checkmark$ & $\checkmark$ & $\checkmark$ & $\checkmark$ & $\checkmark$ & $\circ$ \\
% \midrule
\hline
MPDRL~\cite{WDPH20}         & $\checkmark$ &              & $\checkmark$ & $\checkmark$ & $\checkmark$ & $\ast$ \\
% \midrule
\hline
\textbf{PM-HGNN}            & $\checkmark$ & $\checkmark$ & $\checkmark$ & $\checkmark$ & $\checkmark$ & $\bullet$ \\
% \botrule
\hline
\end{tabular}
% \vspace{-3mm}
% }
\end{table}

In Section~\ref{sec:related_work}, we have systematically discussed related work and highlighted the differences between PM-HGNN and them. 
Here, we further present Table~\ref{table:comparison_diff_models} to summarise the key advantages of PM-HGNN and compares it with a number of recent state-of-the-art methods.
PM-HGNN is the first HGRL model that can adaptively generate personalised meta-paths for each individual node to support node-wise tasks and maintain the end-to-end training mechanism. 

%------------------------------------------------------------------------------

%------------------------------------------------------------------------------
\section{Model Configuration}
\label{sec:appendix_implementation_details}
For the DQN of the proposed PM-HGNN and PM-HGNN\textit{++}, we use the implementation in~\cite{MKSRVBGRF15} with a few modifications to fit it with our frameworks.
We develop a $5$-layers MLP with $(32, 64, 128, 64, 32)$ as the hidden units for $Q$ function.
The memory size is $50 \times b$, where $b$ is the number of validation nodes in the dataset.
For the HGNN module of PM-HGNN and PM-HGNN\textit{++}, we randomly initialise parameters and optimise the model with Adam optimiser.
% ~\cite{KB15}.
We set the learning rate to $0.005$, the regularisation parameter to $0.0001$, the representation vector dimension is $128$, the dimension of the attention vector $Att(\cdot)$ to $16$, the training batch size to $256$ and the number of attention head is $8$, with a dropout ratio to $0.5$.
The max steps ($T$) for IMDb and DBLP are $2$ and $4$, respectively.
For a fair comparison, we set the node representation dimension of all the models mentioned above to $64$.

For random walk-based methods, including DeepWalk, ESim, metapath2vec and HERec, we set the window size to $5$, walk length to $100$, walks per node to $40$ and the number of negative samples to $5$.
For GAT, HAN, and MAGNN, we set the number of attention heads to $8$. 
For HAN and MAGNN, we set the dimension of the attention vector in inter-meta path aggregation to $128$.
For meta-path guided methods, including Esim, metapath2vec, HERec, HAN and MAGNN, we give them the pre-defined meta-paths as in~\cite{WJSWYCY19}. For IMDb, there are two meta-paths: {\it Movie $\rightarrow$ Actor $\rightarrow$ Movie} and {\it Movie $\rightarrow$ Director $\rightarrow$ Movie}. 
For DBLP, there are three meta-paths: {\it Author $\rightarrow$ Paper $\rightarrow$ Author}, 
{\it Author $\rightarrow$ Paper $\rightarrow$ Term $\rightarrow$ Papper $\rightarrow$ Author}, 
{\it Author $\rightarrow$ Paper $\rightarrow$ Venue $\rightarrow$ Papper $\rightarrow$ Author}, and
{\it Venue $\rightarrow$ Paper $\rightarrow$ Author}.
For the relational learning model, we use the implementation published with the official paper and adopt model settings the same as the official settings for the IMDb dataset. 
For relational learning and graph neural network-based models, we test them with the same parameters as PM-HGNN on the same data split.
Competing models are implemented with Pytorch\footnote{\url{https://pytorch.org/}} following the published implementations.

%------------------------------------------------------------------------------

% %------------------------------------------------------------------------------
% \section{More Experimental Results}
% \label{sec:appendix_more_experimental_results}
% \input{pages/appendix_more_experimental_results}
% %------------------------------------------------------------------------------

%------------------------------------------------------------------------------

% \newpage
% \section{Cover letter}
% \input{pages/cover_letter}

% %------------------------------------------------------------------------------
% \clearpage
% \input{pages/response_letter}
% %------------------------------------------------------------------------------

% %------------------------------------------------------------------------------
% \clearpage
% \input{pages/response_letter_2}
% %------------------------------------------------------------------------------

\end{document}